\documentclass{article}

\PassOptionsToPackage{numbers,compress}{natbib}

\usepackage{amsmath}
 \usepackage[preprint]{neurips_2026}

\usepackage[utf8]{inputenc} 
\usepackage[T1]{fontenc}    
\usepackage{hyperref}       
\usepackage{url}            
\usepackage{booktabs}       
\usepackage{amsfonts}       
\usepackage{amssymb}
\usepackage{nicefrac}       
\usepackage{microtype}      
\usepackage{wrapfig}
\usepackage{pifont}
\usepackage{dsfont}
\usepackage{graphicx}
\usepackage{makecell}
\usepackage{enumitem}
\usepackage{multirow}
\usepackage{titletoc}
\usepackage{fontawesome5}

\usepackage[table]{xcolor} 
\newcommand{\xmark}{\ensuremath{\times}}

\usepackage{subcaption}

\title{Same Architecture, Different Capacity:  Optimizer-Induced Spectral Scaling Laws }

\author{
  Nandan Kumar Jha \\
  New York University \\
  \texttt{nj2049@nyu.edu}
  \And
  Brandon Reagen \\
  New York University \\
  \texttt{bjr5@nyu.edu}
}

\begin{document}

\maketitle

\vspace{-3em}
\begin{center}
\textcolor{blue}{\faGlobe{}}: \url{https://optimizer-scaling-laws.github.io}

\end{center}

\begin{abstract}
Scaling laws have made language-model performance predictable from model size,
data, and compute, but they typically treat the optimizer as a fixed training
detail. We show that this assumption misses a fundamental axis of representation
scaling: how effectively the optimizer converts added FFN width into utilized
spectral capacity. Using eigenspectra of feed-forward network representations,
measured through soft and hard spectral-ranks, we find that \emph{the
same Transformer architecture realizes markedly different spectral scaling laws
when trained with different optimizers}. Holding architecture and width schedule
fixed, AdamW exhibits weak hard-rank scaling ($\beta$=0.44) on
rare-token (TAIL) representations where learning is known to be hardest, whereas Muon
achieves linear scaling ($\beta$={\bf 1.02}) in the same regimes, a $2.3\times$ increase in the scaling
exponent. This difference is not reducible to validation loss: AdamW
configurations can match low-rank Dion variants in perplexity, under extended
training, while exhibiting sharply different spectral geometry, demonstrating
that \emph{matched loss does not imply matched representation structure}.
Hard--soft rank asymmetry further reveals that optimizers differ not only in how
much capacity is realized, but also in how that capacity is
structured across eigenmodes. To disentangle optimizer effects from architectural
ones, we compare against architectural interventions (e.g., attention rank and
positional encoding), and find that optimizer-induced
spectral shifts often exceed the architectural effects. These results suggest optimization as a first-class axis of representation scaling, motivating optimizer--architecture co-design.
\end{abstract}

\section{Introduction}
\label{sec:introduction}

Classical scaling-law studies showed that language-model loss follows
predictable power-law trends with model size, training data, and
compute~\citep{kaplan2020scaling,hoffmann2022chinchilla}. This
resource-centric view has made scaling actionable: given a compute budget, one
can estimate how to allocate resources across parameters and data. Yet a central
component of training remains largely outside this framework---the optimizer.
Growing evidence suggests that optimizers do more than affect convergence speed;
they also shape the representations learned by a model through implicit
inductive biases~\citep{pascanu2025optimizers,bernstein2024old}.

Recent work has begun to incorporate optimizer effects into loss-level scaling.
\citep{volkova2026towards} show that loss-scaling exponents can remain shared
across optimizers while multiplicative efficiency factors differ. This provides
a useful abstraction for predicting validation loss, but it leaves open a
representation-level question: \emph{if two optimizers exhibit similar loss
scaling, do they also learn similar internal representations?} If matched loss
can arise from different representation geometries, then optimizer choice is not
merely an efficiency knob. It is a design axis that determines how model
capacity is structured across eigenmodes, allocated across token regimes, and
realized during training.

\begin{figure}[t]
\centering
\subfloat{\includegraphics[width=.49\textwidth]{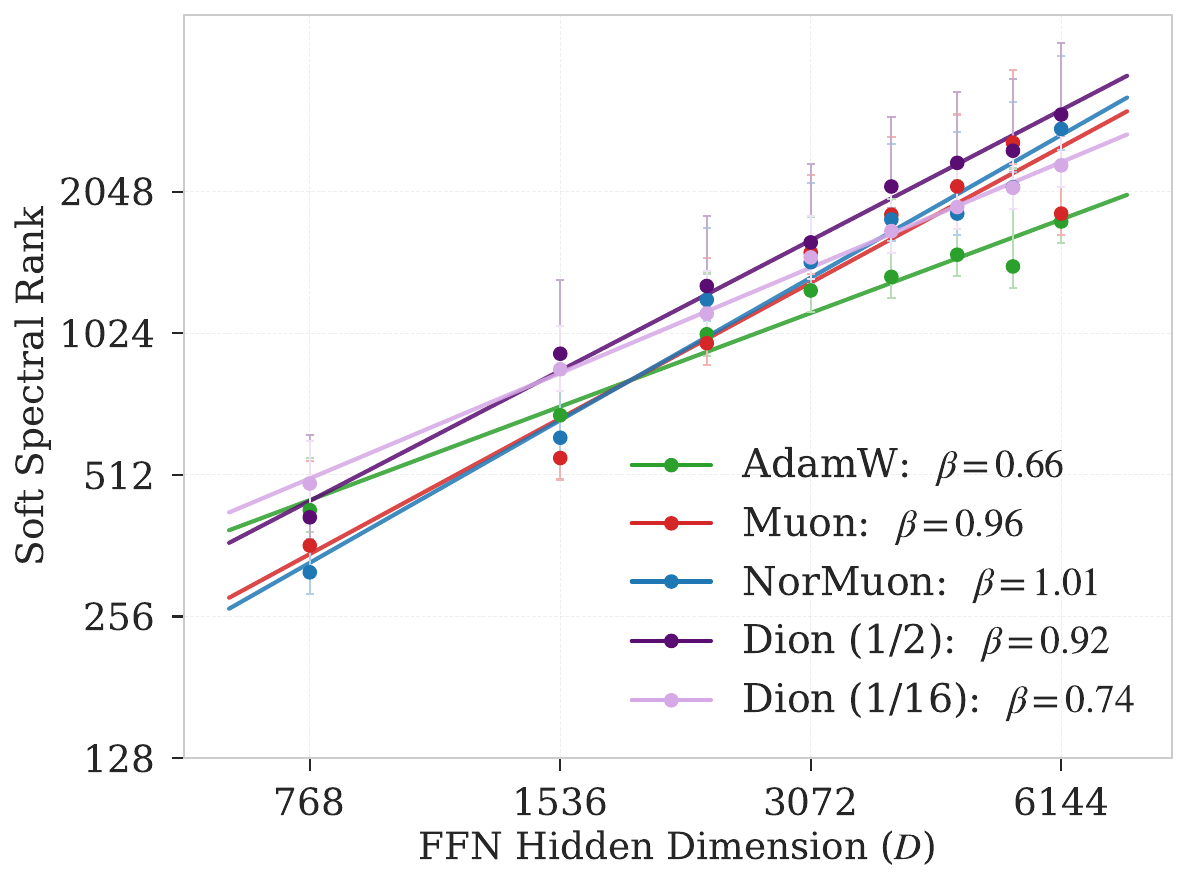}} 
\subfloat{\includegraphics[width=.49\textwidth]{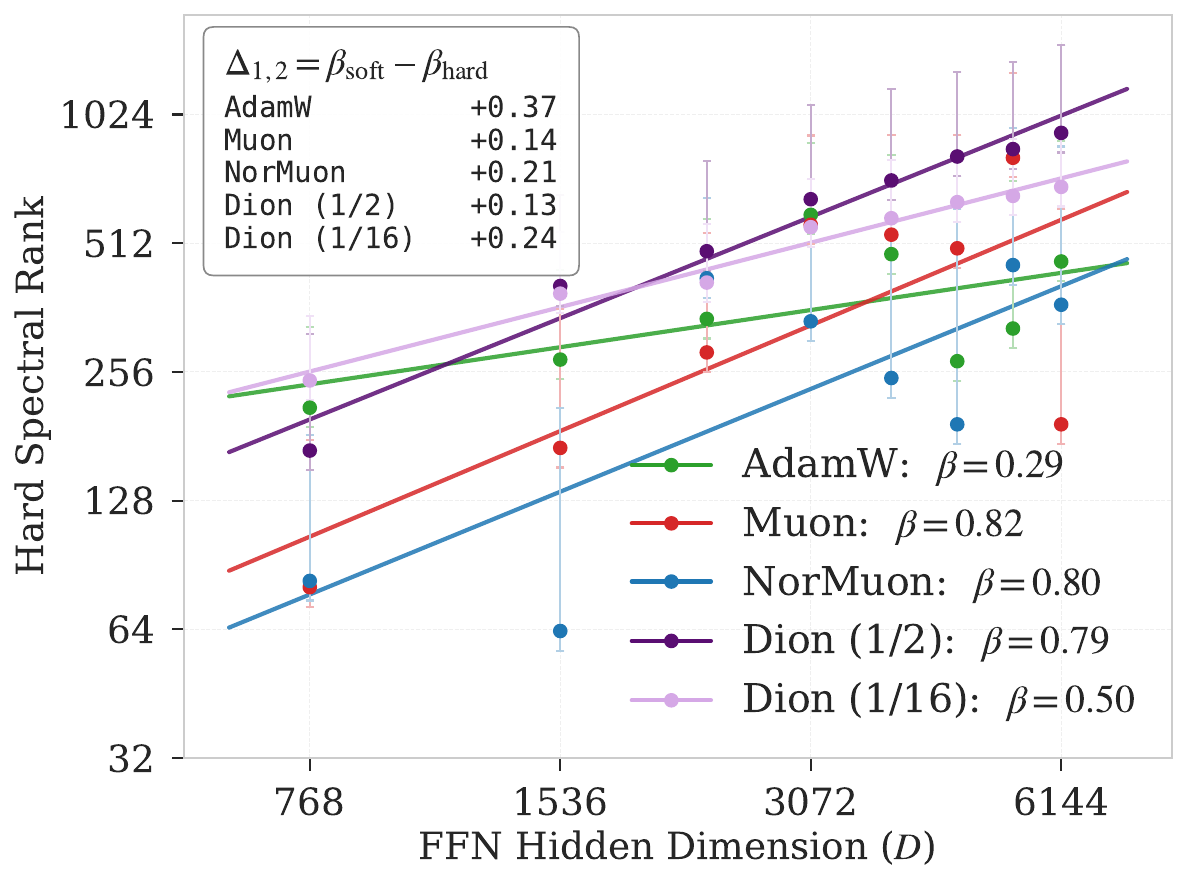}} \vspace{-0.6em}
\caption{Spectral scaling exponents depend on optimizer choice: Soft (left) and hard spectral rank (right) as a function of FFN width in GPT-2 160M. AdamW exhibits the largest hard-soft asymmetry ($\Delta_{1,2} = 0.37$), indicating concentrated eigenspectra. Muon and Dion (1/2) reduce this asymmetry to $\Delta_{1,2} \approx 0.14$.  Moreover, hard-rank scaling exhibits stronger dependence on optimizer choices.}
\label{fig:IntroHardSoftRanks}
\end{figure}


We study this question through the eigenspectra of feed-forward network (FFN)
representations~\citep{jha2026nerve}. FFNs provide a natural setting for this
analysis: in standard Transformer architectures, they account for roughly
two-thirds of model parameters~\citep{geva2020transformer}, and their
expansion--nonlinearity--compression structure exposes a spectrally measurable
latent space. This structure allows us to ask how efficiently added FFN width is
converted into utilized spectral capacity. Building on prior
work~\citep{jha2025spectral}, which established spectral scaling laws for FFN
utilization under a fixed optimizer, we ask whether these laws are invariant to
optimizer choice or instead depend on the architecture--optimizer pair. We
compare AdamW~\citep{loshchilov2018decoupled},
Muon~\citep{jordan2024muon,liu2025muon}, NorMuon~\citep{li2025normuon}, and
rank-constrained Dion variants~\citep{ahn2025dion}, measuring how spectral
capacity is realized during training and how it scales with FFN width.

Holding the architecture and width schedule fixed, changing the optimizer alone
yields markedly different spectral scaling laws.
Figure~\ref{fig:IntroHardSoftRanks} illustrates this phenomenon. We measure FFN
capacity using soft spectral rank, which captures entropy-weighted spectral
spread, and hard spectral rank, which is more sensitive to concentration in
dominant eigenmodes~\citep{jha2025spectral}. The two metrics respond
differently to optimizer choice: soft rank grows substantially with width across
all optimizers, with scaling exponents clustered in a narrow range
$\beta_{\text{soft}} \in [0.66, 1.01]$, whereas hard rank is strongly
optimizer-dependent, spanning $\beta_{\text{hard}} \in [0.29, 0.82]$. AdamW
exhibits weak hard-rank scaling ($\beta$=0.29), while Muon achieves near-linear
scaling ($\beta$=0.82), under identical
architecture  and training data.

This result changes how spectral scaling should be interpreted. In prior work,
the gap between soft-rank and hard-rank scaling appeared to be a stable property
of the Pre-LN Transformer architecture~\citep{jha2025spectral}. We show that
this hard--soft asymmetry is itself optimizer-dependent. AdamW exhibits the
largest asymmetry ($\Delta_{1,2}=0.37$), while Muon-style optimizers
substantially reduce this gap; in particular, Muon and Dion($1/2$) both reach
$\Delta_{1,2}\approx 0.14$. Hence, added FFN width is not automatically
converted into usable capacity. The optimizer helps determine whether extra
dimensions become dominant representational directions or remain diffuse
spectral mass.

This spectral divergence is not apparent from validation loss alone. AdamW
configurations can match matrix-aware optimizer variants in perplexity under
extended training, while their representation spectra remain structurally
distinct. Thus, \emph{matched loss does not imply matched representation
scaling}, and neither learning-rate tuning nor extended training closes this gap.
Therefore, optimizer shapes the geometry of the learned representation, not
just the convergence speed.

The optimizer effect is also structured across the token distribution. Language
data follows a Zipfian distribution, and LLMs are known to struggle
disproportionately with rare and long-tail knowledge~\citep{kandpal2023large}.
We therefore stratify token representations by frequency and measure spectral
scaling separately across frequency regimes. AdamW exhibits especially weak
hard-capacity scaling on rare-token representations, while the largest
AdamW-to-Muon scaling gain appears in the mid-frequency regime. This shows that
optimizer geometry changes not only aggregate representation capacity, but also
how capacity is allocated across the token-frequency distribution.


Finally, we compare optimizer-induced effects against architectural
interventions in attention rank~\citep{amsel2025quality} and positional
encoding~\citep{su2024roformer,chi2023latent}. Optimizer-induced spectral shifts often
dominate or reshape these architectural effects, increasing per-head attention rank
produces smaller spectral changes than switching optimizers, RoPE removal yields
optimizer-dependent redistribution, and orthonormal optimizers enable partial
PostLN configurations to reach useful perplexity where AdamW fails. These
results show that architectural capacity is not realized independently of
optimization, architectural changes are expressed through optimizer geometry.

\vspace{0.3em}
\noindent\textbf{Contributions.} Our contributions are as follows:
\begin{enumerate}[noitemsep,nolistsep,leftmargin=0.5cm] \vspace{-0.4em}

    \item \textbf{Optimizer-induced spectral scaling laws.}
    We show that the same Transformer architecture realizes substantially
    different FFN spectral-capacity scaling laws depending on the optimizer.
    Hard--soft rank asymmetry is optimizer-dependent, exposing spectral scaling
    as a property of the architecture--optimizer pair rather than architecture alone. 
    R\'enyi-entropy spectral analysis further confirms that the
optimizer-induced differences persist across concentration regimes.

    \item \textbf{Matched loss $\neq$ matched geometry.}
    We demonstrate that optimizer-induced spectral differences are not explained
    by learning-rate tuning, convergence speed, or final validation loss:
    configurations with matched perplexity can exhibit distinct spectral
    geometries.

    \item \textbf{Frequency- and update-rank-dependent capacity allocation.}
    We show that optimizer-induced spectral scaling varies across
    token-frequency regimes, with mid- and low-frequency tokens showing the
    strongest effects. Dion rank further acts as a control knob for
    hard-capacity growth.


    \item \textbf{Optimizer--architecture co-design.}
    We show that optimizer-induced spectral shifts can exceed or reshape the
    effects of architectural interventions, motivating joint
    optimizer--architecture design.   
    
\end{enumerate}

\section{Related Work}
\label{sec:related}

\paragraph{Scaling laws and optimizer-aware scaling.}
Classical scaling laws established predictable power-law relationships between
validation loss, model size, data, and compute~\citep{kaplan2020scaling,
hoffmann2022chinchilla}, while treating optimizer as a fixed
training choice~\citep{loshchilov2018decoupled,kingma2014adam,
shazeer2018adafactor}. Recent work shows that scaling behavior can also depend
on inductive biases beyond raw resources: \citep{ngo2026scaling} find
\emph{architecture-dependent exponents} in neural force fields, where
equivariant architectures achieve more favorable power-law slopes. Further, \citep{volkova2026towards} propose optimizer-aware loss
scaling for LLM pretraining, modeling optimizer differences as multiplicative
efficiency factors on shared exponents across AdamW, Muon,
Shampoo~\citep{gupta2018shampoo}, SOAP~\citep{vyas2025soap}, and
Scion~\citep{pethick2025training}. This provides a useful abstraction in which
optimizers act primarily as efficiency rescalings for loss-level prediction. In
contrast, we ask whether optimizer choice changes the scaling exponents of
learned representation, even when validation perplexity is
matched.

\paragraph{Spectral capacity and effective rank.}
Spectral measures have been used to characterize the effective dimensionality
and utilization of learned representations~\citep{roy2007effective,
garrido2023rankme,wei2024diff,skean2025layer}. Moving from loss to representation scaling,
\citep{jha2025spectral} introduced spectral scaling laws for FFN latent-space
utilization in LLMs under a fixed optimizer. In a complementary direction,
\citep{nakis2026rank} showed that utilized capacity is different from the nominal capacity in graph models. Our work studies a different regime and demonstrate that the realized spectral capacity is itself optimizer-dependent.

\paragraph{Optimizer geometry.}
Recent optimizer work suggests that training algorithms impose nontrivial
geometry on matrix-valued parameters. Muon~\citep{jordan2024muon}
orthogonalizes matrix updates via Newton--Schulz
iterations~\citep{kim2026convergence,amsel2026the}. Large-scale studies further
show that Muon-style and other matrix-based optimizers can be competitive for
LLM pretraining, although their gains depend on tuning, scale, and evaluation
protocol~\citep{liu2025muon,wen2026fantastic}. Dion~\citep{ahn2025dion}
provides a particularly useful intervention for our analysis since its
rank-constrained orthonormalized updates allow us to separate update geometry
from update rank. Existing studies focus primarily on optimizer mechanisms and
training efficiency.


\paragraph{Architecture--optimizer interaction.}
\citep{ding2023repoptimizer} showed that architectural
priors can be transferred into optimizers through gradient reparameterization,
folding inductive biases that would normally live in the architecture into the
optimizer update rule. PoLAR~\citep{lion2025polar} provides another example of
architecture--optimizer co-design, pairing structured low-rank parameterization
with Riemannian optimization on the manifold induced by the parameterization.
NerVE~\citep{jha2026nerve} showed that
optimizer geometry modulates how nonlinearities redistribute variance within
fixed-width FFNs. We extend this analysis to the scaling regime, showing that
optimizer choice systematically changes the scaling \emph{exponents} of FFN
representation. We further show that architectural interventions do not induce
optimizer-independent spectral shifts, their effects can be exceeded or
reshaped by optimizer geometry.
\section{Methodology}
\label{sec:method}

We measure how optimizer choice changes the effective latent capacity of FFNs
under a fixed Transformer architecture. Three choices are central to the
measurement. First, capacity is not one-dimensional: effective ranks with
different concentration sensitivities can scale differently with FFN width. We therefore work within the R\'enyi effective-rank
family rather than relying on a single rank estimate
(Table~\ref{tab:renyi_alpha_interpretation}). Second, probe location
matters. Pre-activation states capture optimizer-induced geometry before the FFN
nonlinearity, while post-activation states capture the capacity realized after
nonlinear transformation. Third, aggregate spectra can hide token-frequency-dependent effects; 
hence, we stratify FFN representations by token frequency. Throughout,
\emph{spectral capacity} denotes the effective dimensionality of
variance-bearing directions in the FFN latent space.

\subsection{FFN probe points and covariance spectra}
\label{subsec:probe_points}

For layer $\ell$, we probe the FFN at two complementary states: the
pre-activation state $z_{\ell,t}$ and the post-activation state $a_{\ell,t}$:
\begin{equation}
\label{eq:probe_states}
z_{\ell,t} = W_{\mathrm{in}}^{(\ell)} x_{\ell,t},
\qquad
a_{\ell,t} = \phi(z_{\ell,t}).
\end{equation}
The two probes answer complementary questions. Pre-activation spectra expose the
optimizer-shaped linear geometry before the FFN nonlinearity, whereas
post-activation spectra measure the realized latent capacity passed to the output
projection. Their comparison gives a three-stage view of FFN capacity: the
optimizer shapes the linear expansion, the nonlinearity redistributes spectral
mass, and the post-activation state determines the capacity available to
subsequent layers.

Given FFN representations $h_n \in \mathbb{R}^D$ from either probe point, we
compute the empirical covariance and trace-normalized eigenspectrum: \vspace{-0.6em}
\begin{equation}
\label{eq:covariance_eigenspectrum}
C = \frac{1}{N-1}\sum_{n=1}^{N}(h_n-\mu)(h_n-\mu)^\top,
\quad
\mu = \frac{1}{N}\sum_{n=1}^{N}h_n,
\quad
p_i = \frac{\lambda_i(C)}{\sum_{j=1}^{D}\lambda_j(C)} .
\end{equation}
The distribution $p$ describes how variance is allocated
across FFN latent directions, with $p_i \geq 0$ and $\sum_i p_i = 1$.
Trace normalization makes spectra comparable across probe points, layers, and
widths.

\subsection{R\'enyi-family effective rank}
\label{subsec:renyi_rank}

\paragraph{R\'enyi entropy and the effective-rank.}
Using the normalized eigenspectrum $p$, we quantify spectral spread through
the R\'enyi entropy family~\citep{renyi1961measures,van2014renyi}. The order
$\alpha$ controls concentration sensitivity: lower orders give more weight to
weak eigendirections and diffuse spectral support, while higher orders
increasingly emphasize dominant eigendirections. We define
\begin{equation}
\label{eq:renyi_entropy}
H_\alpha(p)=
\begin{cases}
\frac{1}{1-\alpha}\log\sum_{i=1}^{D}p_i^\alpha,
& \alpha>0,\ \alpha\neq 1,\\[2pt]
-\sum_{i=1}^{D}p_i\log p_i,
& \alpha\to 1,
\end{cases}
\qquad
R_\alpha(p)=\exp(H_\alpha(p)).
\end{equation}
For $\alpha\neq1$, this gives
$R_\alpha(p)=\left(\sum_i p_i^\alpha\right)^{1/(1-\alpha)}$. Thus,
$R_\alpha$ defines a continuum of effective-rank measures with different
concentration sensitivities, unifying entropy-based effective
rank~\citep{roy2007effective,garrido2023rankme,wei2024diff} and
participation-ratio effective
dimension~\citep{gao2017theory,hu2022spectrum,xie2025slow,ruscio2025what}
within a single information-theoretic framework. Table~\ref{tab:renyi_alpha_interpretation}
summarizes how different $\alpha$ values probe different aspects of the
spectrum.

\paragraph{Soft and hard rank as primary anchors.}
For the main scaling-law analyses, we anchor at $\alpha=1$ and $\alpha=2$,
which correspond to two standard notions of effective rank:
\begin{equation}
\label{eq:soft_hard_rank}
R_1(p)=\exp\!\Bigl(-\sum_i p_i\log p_i\Bigr)
\quad\text{(soft rank)},
\qquad
R_2(p)=\frac{1}{\sum_i p_i^2}
=\frac{(\sum_i\lambda_i)^2}{\sum_i\lambda_i^2}
\quad\text{(hard rank)}.
\end{equation}
The soft rank $R_1$ measures Shannon-like entropy-weighted spectral spread,
while the hard rank $R_2$ is the participation ratio and provides a stricter,
concentration-sensitive measure of effective dimensionality. These two anchors
capture complementary aspects of spectral capacity: $R_1$ is sensitive to
diffuse spread across many directions, whereas $R_2$ is more strongly affected
by dominant eigendirections. The full R\'enyi sweep across
$\alpha\in\{0.5,1,1.5,2,3,5\}$ is reported in
Appendix~\ref{sec:renyi_family} and tests whether optimizer-induced capacity
differences persist across concentration regimes.

\begin{table}[t]
\centering
\caption{Interpretation of Rényi order $\alpha$ for normalized FFN eigenspectra.}
\label{tab:renyi_alpha_interpretation}
\resizebox{0.99\textwidth}{!}{%
\begin{tabular}{lll}
\toprule
\textbf{Order} & \textbf{Sensitivity} & \textbf{Spectral interpretation} \\
\midrule
$0<\alpha<1$ 
& Weak-direction-sensitive 
& Gives more credit to small nonzero eigenvalues \\

$\alpha=1$ 
& Shannon-balanced 
& Measures entropy-weighted spectral spread \\

$1<\alpha<2$ 
& Mildly concentration-sensitive 
& Interpolates Shannon spread and order-2 concentration \\

$\alpha=2$ 
& Quadratic / participation-ratio regime 
& Captures concentration-sensitive effective rank \\

$\alpha>2$ 
& Strongly concentration-sensitive 
& Increasingly dominated by leading eigendirections \\
\bottomrule
\end{tabular}}
\end{table}

\paragraph{Hard--soft asymmetry.}
Since R\'enyi entropy is non-increasing in $\alpha$, $R_1(p)\geq R_2(p)$
for any spectrum~\citep{van2014renyi}. We define the rank-level hard--soft
asymmetry as
\begin{equation}
\label{eq:rank_level_asymmetry}
A_{1,2}(p)=\log R_1(p)-\log R_2(p)\geq 0,
\end{equation}
with larger values indicating more concentrated eigenspectra. For scaling-law
fits, we report the corresponding exponent-level asymmetry
\begin{equation}
\label{eq:asymmetry}
\Delta_{1,2}=\beta_{\mathrm{soft}}-\beta_{\mathrm{hard}},
\end{equation}
where $\beta_{\mathrm{soft}}$ and $\beta_{\mathrm{hard}}$ are obtained by
fitting $R_1(D)$ and $R_2(D)$ as functions of FFN width $D$. Higher asymmetry 
indicates that added width expands low-variance directions more than the dominant ones.


\subsection{Token-frequency stratification}
\label{subsec:freq_stratification}

Aggregate spectra can be dominated by frequent tokens and may obscure
the capacity scaling for rarer tokens. Motivated by the long-tailed structure of
language data, we stratify FFN representations by token frequency, 
where frequency regimes are defined over token \emph{types}, rather than knowledge concepts
or factual entities. This connects most directly to token-level analyses
of frequency-dependent scaling~\citep{kunstner2025scaling}, while the broader
difficulty of rare and long-tail knowledge in language models is supported
by~\citep{kandpal2023large}. Let $f(v)$ be the corpus frequency of
token type $v\in\mathcal{V}$ and $M=\sum_{v\in\mathcal{V}} f(v)$ be the total
occurrence. We sort token types by decreasing frequency and choose
thresholds $\tau_{\mathrm{head}}$ and $\tau_{\mathrm{mid}}$ so that the top
regime covers approximately one third of occurrences and the top two regimes
cover approximately two thirds:
\begin{equation}
\label{eq:freq_stratification}
\sum_{v:f(v)\ge \tau_{\mathrm{head}}} f(v) \approx \frac{M}{3},
\;
\sum_{v:f(v)\ge \tau_{\mathrm{mid}}} f(v) \approx \frac{2M}{3},
\;
b(v)=
\begin{cases}
\mathrm{HEAD}, & f(v)\ge \tau_{\mathrm{head}},\\
\mathrm{MID}, & \tau_{\mathrm{mid}}\le f(v)<\tau_{\mathrm{head}},\\
\mathrm{TAIL}, & f(v)<\tau_{\mathrm{mid}}.
\end{cases}
\end{equation}
HEAD contains the most frequent token types covering the top third of
occurrence mass, MID covers the next third, and TAIL contains the remaining
lower-frequency types. For each regime $b$, we compute covariance spectra and
rank metrics on $\mathcal{H}_b=\{h_n:b(v_n)=b\}$. This stratification allows us to understand how
optimizer-induced spectral-capacity varies across the token
distribution.

\subsection{Scaling laws for effective FFN capacity}
\label{subsec:scaling_laws} 

We vary the FFN hidden dimension as $D=md_{\text{model}}$ with $m\in\mathbb{Z}$ and
compute post-activation soft and hard ranks for each layer, across
token-frequency regime (HEAD, MID, TAIL). We fit 
\begin{equation}
\label{eq:scaling_law}
R(D)\propto D^\beta
\qquad\Longleftrightarrow\qquad
\log R(D)=\beta\log D+c,
\end{equation}
where $R\in\{R_1,R_2\}$ is computed on aggregate or frequency-stratified spectra.
The same machinery applies to any R\'enyi order $\alpha$, allowing us to fit a
scaling exponent $\beta_\alpha$ at different concentration sensitivities. The
main analyses focus on $\alpha=1$ and $\alpha=2$, with the broader $\alpha$
sweep reported in Appendix~\ref{sec:renyi_family}. The exponent $\beta$
measures how efficiently added FFN width is converted into effective latent
capacity in the probed concentration regime.

\section{Optimizer-Induced Spectral Scaling Laws}

\paragraph{Experimental setup}
We train GPT-style decoder-only Transformers on
FineWeb-Edu~\citep{penedo2024fineweb} following the
modded-nanoGPT~\citep{modded_nanogpt_2024} configuration: Pre-RMSNorm,
RoPE~\citep{su2024roformer}, squared-ReLU~\citep{so2021searching} FFNs, no
biases, and QK-normalization \citep{dehghani2023scaling}. Our primary experiments use 160M base models, and we replicate the main
scaling trends on 350M base models. The 160M and 350M labels
denote the base ($4\, d_{\text{model}}$) configuration; we vary the FFN hidden dimension
as $D = m\, d_{\text{model}}$ with $m \in \{1, \ldots, 8\}$ at 160M and
$m \in \{1, 2, 3, 4\}$ at 350M, hence total parameter count grows with FFN width across the sweep.
We compare AdamW~\citep{loshchilov2018decoupled}, Muon~\citep{jordan2024muon},
NorMuon~\citep{li2025normuon}, and Dion~\citep{ahn2025dion} at rank fractions
$r \in \{1/2, 1/4, 1/8, 1/16\}$. The 160M variants are trained on 3.15B tokens, 
the 350M variants on 4.19B tokens, both with sequence length of 512 and global batch size 1024.  
Full training configurations, optimizer and training
hyperparameters are deferred to
Appendix~\ref{app:experimental_setup}.


\subsection{Spectral Scaling Laws Are Optimizer-Dependent}
\label{sec:results:scaling}

\begin{figure}[htbp]
    \centering
    \includegraphics[width=0.98\textwidth]{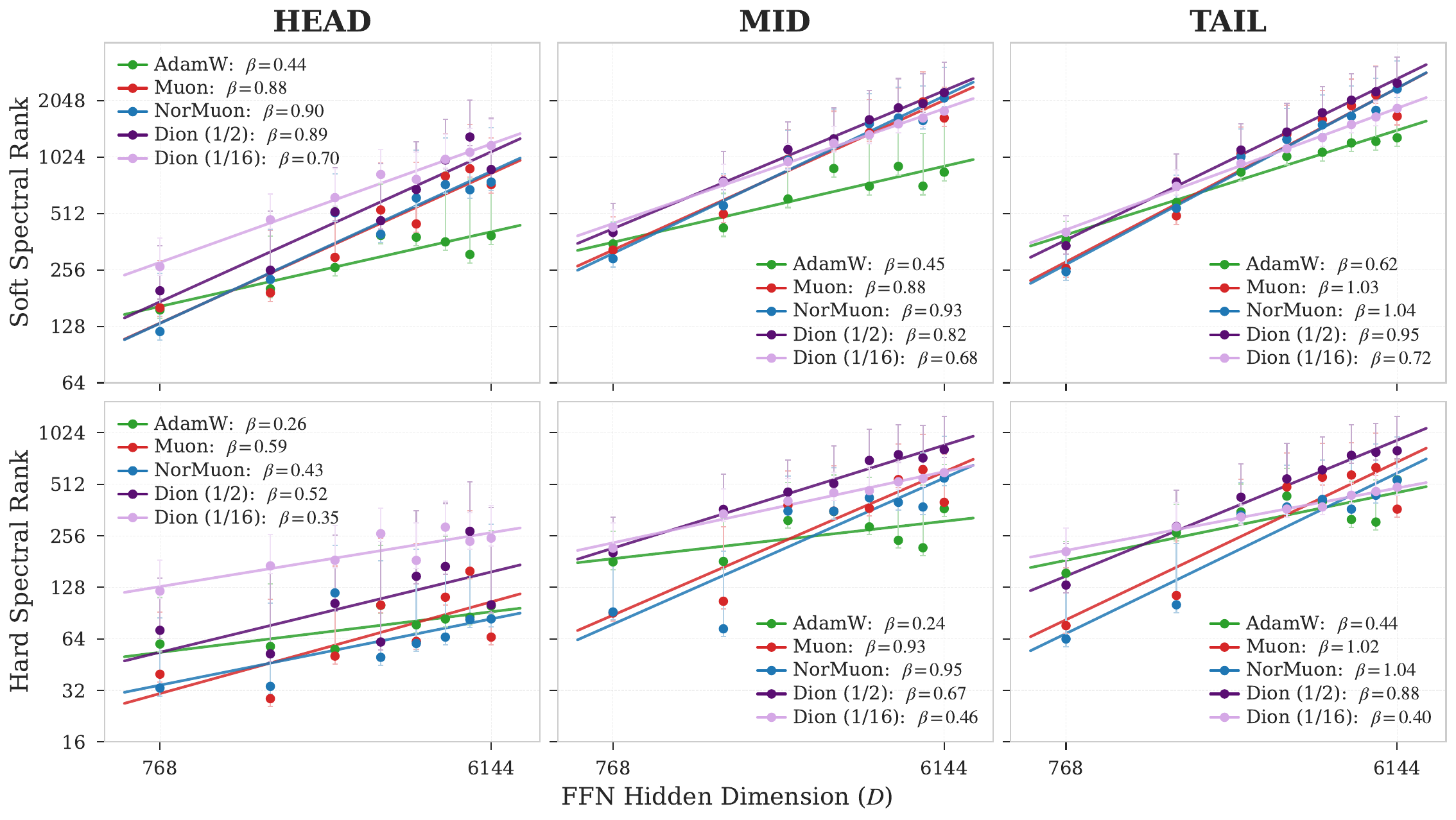} \vspace{-0.6em}
   \caption{Optimizer-dependent spectral scaling across token-frequency regimes.
Soft spectral rank (top) and hard spectral rank (bottom) are shown as functions
of FFN hidden width for HEAD, MID, and TAIL tokens in GPT-2 160M. AdamW exhibits persistent positive hard--soft asymmetry, indicating that
added width contributes mostly to diffuse spectral capacity rather than
dominant-mode capacity. Muon and NorMuon reduce this asymmetry to near zero for MID and TAIL tokens,
whereas low-rank Dion ($r=1/16$) retains AdamW-like asymmetry despite
orthonormalized updates.}
    \label{fig:frequency_scaling_160m}
\end{figure}

We perform spectral scaling analysis separately across
token-frequency regimes to see how capacity is allocated for frequent vs. rare tokens, 
and whether the  optimizer effects are concentrated in particular frequency regimes.
Figure~\ref{fig:frequency_scaling_160m} depicts  the scaling trends, and Table~\ref{tab:architectural_beta_160m}
reports their numerical values. 

\paragraph{Hard-rank scaling is strongly optimizer-dependent.}
Optimizer choice strongly affects hard spectral rank scaling, which measures the
growth of dominant-mode capacity. In the TAIL regime, AdamW reaches
only $\beta_{\mathrm{hard}}=0.44$, whereas Muon and NorMuon achieve 
linear scaling with $\beta_{\mathrm{hard}}=1.02$ and $1.04$, respectively. The
separation is also large in the MID regime: AdamW obtains
$\beta_{\mathrm{hard}}=0.24$, compared with $0.93$ for Muon and $0.95$ for
NorMuon. HEAD tokens show weaker and less reliable hard-rank separation, with
$\beta_{\mathrm{hard}}\in[0.26,0.59]$ and lower fit quality. Thus, MID and TAIL tokens are the most diagnostic regimes for
optimizer-induced scaling effects.

\begin{table}[htbp]
\caption{$\beta$ values for soft and hard ranks with $R^2$ in parentheses for GPT-2 160M (Figure \ref{fig:frequency_scaling_160m}). 
    Positive  $\Delta_{1,2}$ indicates concentrated eigenspectra while lower values 
    indicates better utilization of FFN width. Muon and NorMuon achieve near-zero $\Delta_{1,2}$
    for MID/TAIL tokens, while AdamW maintains $\Delta_{1,2} \approx +0.2$ across all the  token-frequency regimes. }
\label{tab:architectural_beta_160m}
\centering
\resizebox{\textwidth}{!}{%
\begin{tabular}{l ccc ccc ccc}
\toprule
 & \multicolumn{3}{c}{HEAD}
 & \multicolumn{3}{c}{MID}
 & \multicolumn{3}{c}{TAIL} \\
\cmidrule(lr){2-4} \cmidrule(lr){5-7} \cmidrule(lr){8-10}
Optimizer
 & $\beta_{\mathrm{hard}}$ & $\beta_{\mathrm{soft}}$ & $\Delta_{1,2}$
 & $\beta_{\mathrm{hard}}$ & $\beta_{\mathrm{soft}}$ & $\Delta_{1,2}$
 & $\beta_{\mathrm{hard}}$ & $\beta_{\mathrm{soft}}$ & $\Delta_{1,2}$ \\
\midrule
AdamW
 & 0.26 \scriptsize{(0.59)} & 0.44 \scriptsize{(0.82)} & $+$0.18
 & 0.24 \scriptsize{(0.36)} & 0.45 \scriptsize{(0.82)} & $+$0.21
 & 0.44 \scriptsize{(0.66)} & 0.62 \scriptsize{(0.97)} & $+$0.18 \\
Muon
 & 0.59 \scriptsize{(0.54)} & 0.88 \scriptsize{(0.90)} & $+$0.29
 & 0.93 \scriptsize{(0.82)} & 0.88 \scriptsize{(0.96)} & $-$0.04
 & 1.02 \scriptsize{(0.81)} & 1.03 \scriptsize{(0.94)} & $+$0.01 \\
NorMuon
 & 0.43 \scriptsize{(0.45)} & 0.90 \scriptsize{(0.92)} & $+$0.47
 & 0.95 \scriptsize{(0.77)} & 0.93 \scriptsize{(0.98)} & $-$0.02
 & 1.04 \scriptsize{(0.89)} & 1.04 \scriptsize{(0.98)} & $+$0.00 \\
Dion (1/2)
 & 0.52 \scriptsize{(0.43)} & 0.89 \scriptsize{(0.90)} & $+$0.37
 & 0.67 \scriptsize{(0.98)} & 0.82 \scriptsize{(0.99)} & $+$0.15
 & 0.88 \scriptsize{(0.98)} & 0.95 \scriptsize{(0.99)} & $+$0.07 \\
Dion (1/16)
 & 0.35 \scriptsize{(0.75)} & 0.70 \scriptsize{(0.98)} & $+$0.35
 & 0.46 \scriptsize{(0.98)} & 0.68 \scriptsize{(1.00)} & $+$0.22
 & 0.40 \scriptsize{(0.99)} & 0.72 \scriptsize{(1.00)} & $+$0.31 \\
\bottomrule
\end{tabular}%
}
\end{table}

\paragraph{MID tokens show the largest AdamW-to-Muon gain.}
A finer frequency-dependent pattern appears when comparing AdamW and Muon.
Under AdamW, HEAD and MID tokens scale similarly
($\beta_{\mathrm{hard}}=0.26$ and $0.24$), while TAIL tokens scale more
strongly ($\beta_{\mathrm{hard}}=0.44$). Muon changes this structure: MID rises
to $\beta_{\mathrm{hard}}=0.93$, nearly matching TAIL
($\beta_{\mathrm{hard}}=1.02$), while HEAD remains lower
($\beta_{\mathrm{hard}}=0.59$). The AdamW-to-Muon gains are
\[
\Delta\beta_{\mathrm{MID}}=+0.69,\qquad
\Delta\beta_{\mathrm{TAIL}}=+0.58,\qquad
\Delta\beta_{\mathrm{HEAD}}=+0.33 .
\]
Thus, the MID gain is the largest, roughly $2.1\times$ the HEAD gain.

\paragraph{Hard--soft asymmetry reveals how FFN width is utilized.}
AdamW exhibits persistent positive asymmetry across all frequency regimes
($\Delta_{1,2}\approx0.18$--$0.21$), indicating that added width contributes 
mostly to diffuse spectral capacity rather than dominant-mode capacity.
In contrast, Muon and NorMuon nearly eliminate this asymmetry for MID and TAIL
tokens; hence dominant-mode capacity scales at nearly the same rate as
entropy-weighted spectral spread in the critical frequency regimes.

\paragraph{Low-rank optimizer structure constrains hard-capacity scaling.}
Dion separates orthonormalization from update rank. With rank fraction
$r=1/2$, Dion approaches Muon/NorMuon in the TAIL regime, reaching
$\beta_{\mathrm{hard}}=0.88$ with small asymmetry
($\Delta_{1,2}=0.07$). At $r=1/16$, however, TAIL hard-rank scaling drops to
$\beta_{\mathrm{hard}}=0.40$, comparable to AdamW, while the asymmetry rises to
$\Delta_{1,2}=0.31$. Thus, orthonormalization alone is insufficient: the rank
of the optimizer update constrains how efficiently added FFN width becomes
usable hard spectral capacity.

\paragraph{Robustness of the scaling trends.}
Soft-rank fits are consistently strong, indicating that added width reliably
increases entropy-weighted spectral spread. Some HEAD and MID hard-rank fits
have lower $R^2$, so we interpret those exponents as directional evidence of
scaling behavior rather than precise constants. To verify that the aggregate
trends are not artifacts of layer averaging, we also fit layer-wise exponents
$\beta_\ell$ independently for each layer. Appendix
Fig.~\ref{fig:layerwise_beta_distribution_160m} reports the resulting
$\beta_\ell$ distributions, and Appendix
Fig.~\ref{fig:layerwise_beta_depth_160m} shows their depth profiles.


\subsection{Matched Loss Does Not Imply Matched Spectral Geometry}
\label{sec:results:matched_loss}

A plausible explanation for AdamW's lower spectral-scaling exponents is slower
convergence: perhaps longer training is required to match the scaling behavior
of matrix-aware optimizers. We test this convergence-only explanation by
training AdamW for 12K steps and comparing it with Dion($1/16$) at 6K steps,
which achieves similar validation perplexity. This comparison tests whether
matching loss is sufficient to recover the same spectral-capacity scaling.

\paragraph{Extended AdamW training does not recover hard-rank scaling.}
Table~\ref{tab:structural} rejects the convergence-only explanation. Although
AdamW 12K matches Dion~$(1/16)$ in perplexity (Table \ref{tab:appendix_eval_ppl_160m} in Appendix \ref{app:ppl_160m}), its aggregate hard-rank scaling
nearly vanishes: $\beta_{\mathrm{hard}}$ drops from $0.29$ at 6K steps to
$0.03$ at 12K steps. This degradation is also visible in 
HEAD, MID, and TAIL regimes, as their $\beta_{\mathrm{hard}}$
decrease to $0.13$, $0.17$, and $0.18$, respectively. In contrast, Dion~$(1/16)$ maintains
reliable power-law scaling, with $\beta_{\mathrm{hard}}=0.50$ in aggregate and
strong fit quality across frequency regimes ($R^2=0.75$ in HEAD and
$R^2\geq0.98$ in MID/TAIL). Note that the aggregate
soft-rank scaling decreases only mildly, from $\beta_{\mathrm{soft}}=0.66$ to
$0.58$, while hard--soft asymmetry increases from
$\Delta_{1,2}=+0.37$ to $+0.55$. Thus, longer AdamW training fails to convert added FFN width into
dominant-mode capacity.

\begin{table}[htbp]
    \centering
    \caption{Extended AdamW training breaks hard-rank scaling despite matched loss in GPT-2 160M.
    AdamW 12K and Dion~$(1/16)$ 6K achieve similar perplexity
    across width points, but exhibit different spectral geometry:
    AdamW's aggregate hard-rank scaling nearly vanishes
    ($\beta_{\mathrm{hard}}$=0.03, $R^2$=0.01), whereas Dion maintains
    reliable power-law scaling. $R^2$ values are shown in
    parentheses.}
    \label{tab:structural}
    \resizebox{\columnwidth}{!}{%
    \begin{tabular}{l ccc cc cc cc}
    \toprule
    & \multicolumn{3}{c}{Aggregate}
    & \multicolumn{2}{c}{HEAD}
    & \multicolumn{2}{c}{MID}
    & \multicolumn{2}{c}{TAIL} \\
    \cmidrule(lr){2-4}
    \cmidrule(lr){5-6}
    \cmidrule(lr){7-8}
    \cmidrule(lr){9-10}
    Configuration
    & $\beta_{\mathrm{hard}}$
    & $\beta_{\mathrm{soft}}$
    & $\Delta_{1,2}$
    & $\beta_{\mathrm{hard}}$
    & $\Delta_{1,2}$
    & $\beta_{\mathrm{hard}}$
    & $\Delta_{1,2}$
    & $\beta_{\mathrm{hard}}$
    & $\Delta_{1,2}$ \\
    \midrule
    AdamW 6K
    & 0.29 \scriptsize{(0.34)}
    & 0.66 \scriptsize{(0.97)}
    & $+$0.37
    & 0.26 \scriptsize{(0.59)}
    & $+$0.18
    & 0.24 \scriptsize{(0.36)}
    & $+$0.21
    & 0.44 \scriptsize{(0.66)}
    & $+$0.18 \\
    AdamW 12K
    & \textbf{0.03} \scriptsize{(\textbf{0.01})}
    & 0.58 \scriptsize{(0.91)}
    & $\mathbf{+0.55}$
    & 0.13 \scriptsize{(\textbf{0.12})}
    & $\mathbf{+0.28}$
    & 0.17 \scriptsize{(0.42)}
    & $\mathbf{+0.30}$
    & 0.18 \scriptsize{(\textbf{0.29})}
    & $\mathbf{+0.35}$ \\
    Dion (1/16) 6K
    & 0.50 \scriptsize{(0.97)}
    & 0.74 \scriptsize{(1.00)}
    & $+$0.24
    & 0.35 \scriptsize{(0.75)}
    & $+$0.35
    & 0.46 \scriptsize{(0.98)}
    & $+$0.22
    & 0.40 \scriptsize{(0.99)}
    & $+$0.31 \\
    \bottomrule
    \end{tabular}%
    }
\end{table}

\begin{figure}[htbp]
    \centering
    \includegraphics[width=\linewidth]{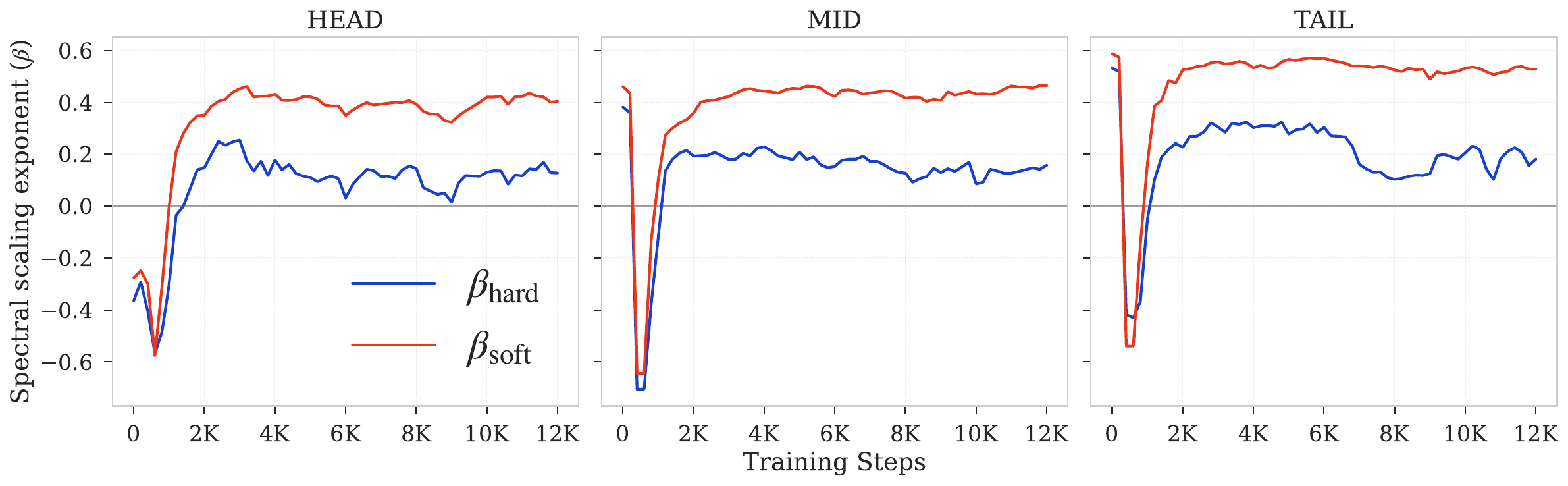} \vspace{-2em}
    \caption{Extended AdamW training weakens hard-rank scaling in GPT-2
160M. Hard-rank scaling ($\beta_{\mathrm{hard}}$) exhibits diminishing return for TAIL token scaling. Soft-rank scaling remains comparatively stable,
leading to increasing hard--soft asymmetry at 12K (see Table~\ref{tab:structural}).}
    \label{fig:beta_dynamics}
\end{figure}

\paragraph{Hard-rank scaling breaks dynamically during training.}
Figure~\ref{fig:beta_dynamics} shows the diminishing return for hard-rank scaling in TAIL regime. TAIL
hard-rank scaling initially improves, peaks $\sim$4K steps, and then declines. 
HEAD degrades more steadily after the early transient, while MID shows
intermediate degradation. Soft-rank scaling remains comparatively stable
throughout, leads to growing hard--soft asymmetry in every frequency regime,
including TAIL ($+0.18\rightarrow+0.35$). The underlying participation-ratio
trajectories explain why the hard-rank power law fails: larger FFN widths lose
PR capacity faster, breaking the monotonic width--capacity ordering required
for a clean scaling law (Appendix~\ref{app:adamw_pr_dynamics}). This show that optimizer-induced spectral scaling is not a transient
convergence artifact, and matched perplexity does not
imply matched spectral capacity: optimizer choice determines whether added FFN
width is converted into systematic hard-rank capacity.

We also show that AdamW--Muon spectral scaling gap is not closed by learning-rate tuning.
Across the learning-rate sweep, Muon's lowest TAIL hard-rank exponent
remains above AdamW's highest valid exponent, indicating that the gap is not a
simple hyperparameter artifact (Appendix~\ref{app:lr_ablation}).


\subsection{Update Rank Constrains Hard-Capacity Scaling}
\label{subsec:dion_rank_sweep}

We next isolate which component of orthonormalized-update optimizers drives
spectral capacity growth. Dion provides a controlled intervention: varying the
rank fraction $r$ changes the rank of the projected update while preserving the
orthonormalized-update structure. This lets us test whether update rank controls
how added FFN width is converted into soft and hard spectral capacity.

\begin{figure} [htbp]
\centering
\subfloat{\includegraphics[width=.49\textwidth]{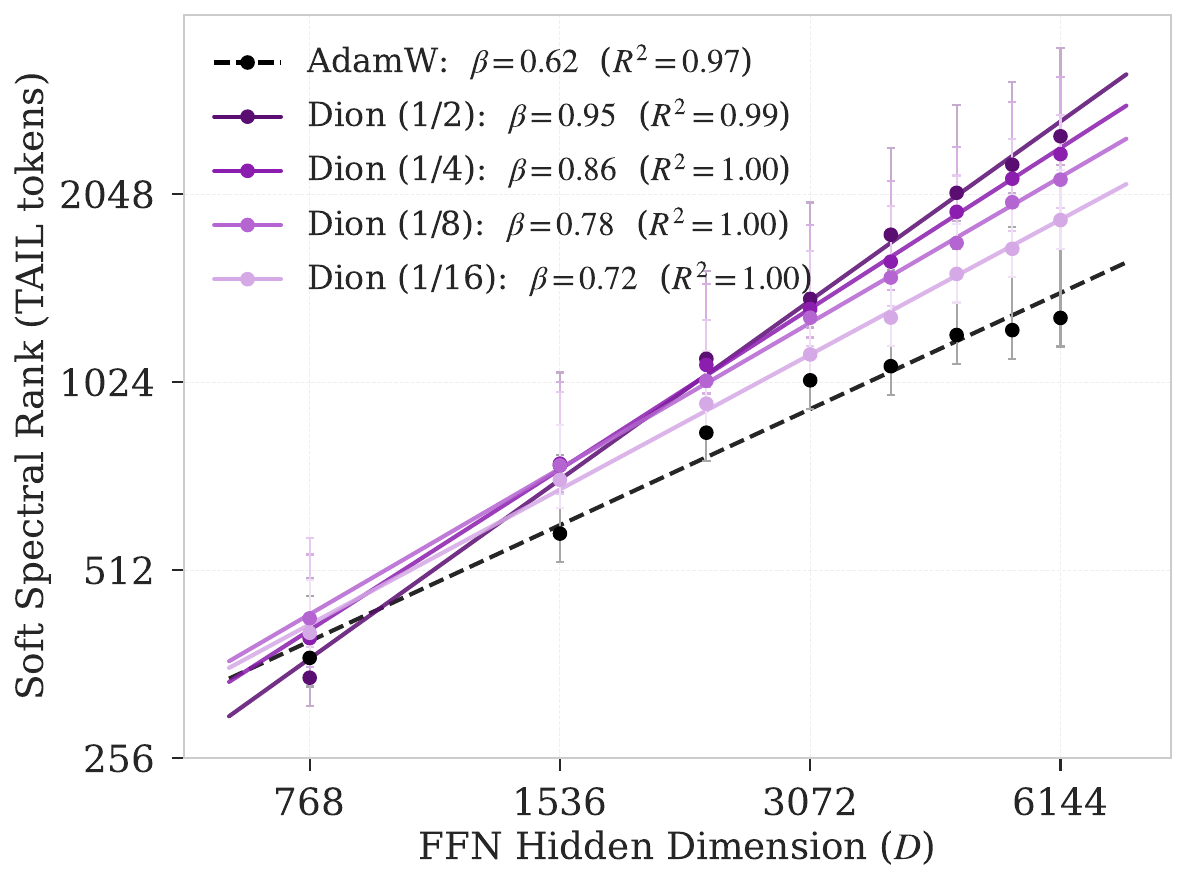}} 
\subfloat{\includegraphics[width=.49\textwidth]{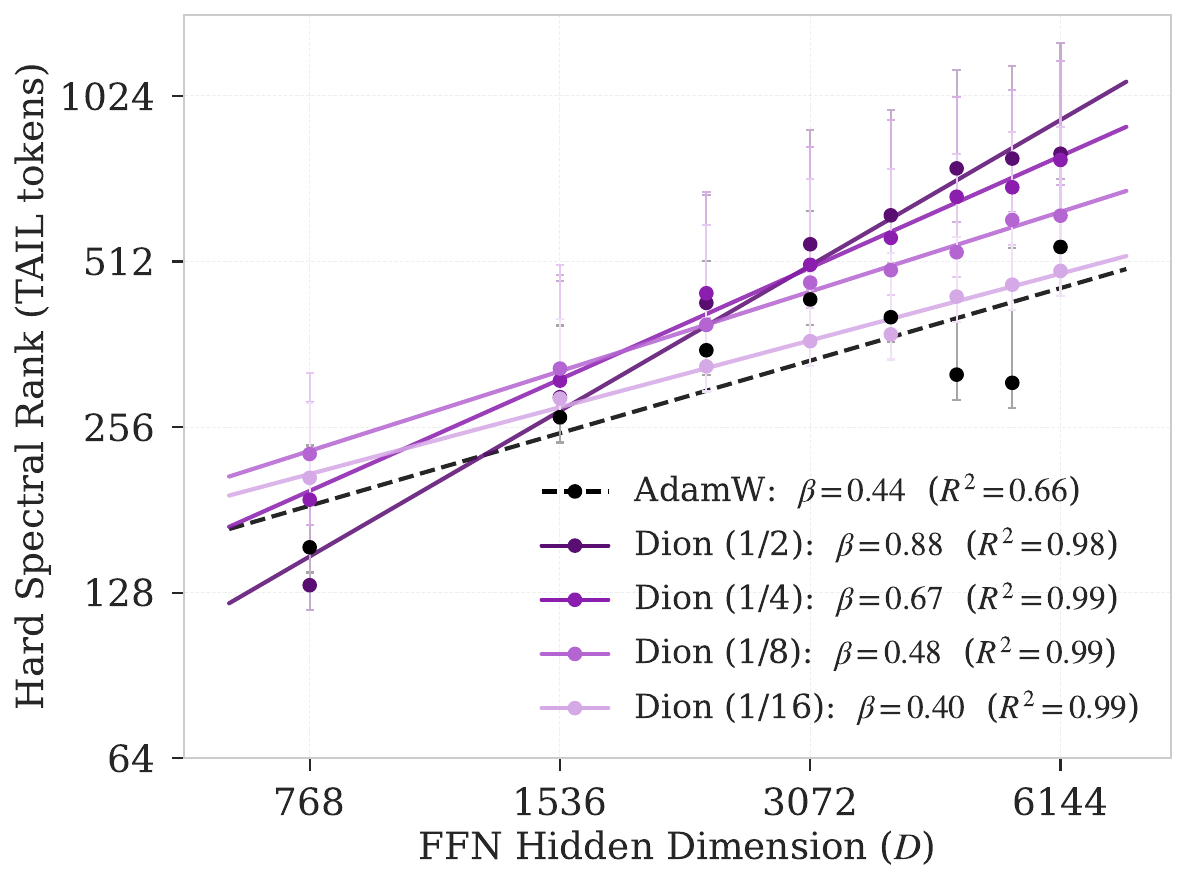}} \vspace{-0.5em}
\caption{Soft spectral rank (left) and hard spectral rank (right) scaling is shown for TAIL tokens in GPT-2 160M, with AdamW as a reference. As the
Dion rank fraction \(r\) decreases, hard-rank scaling drops from
\(\beta=0.88\) at \(r=1/2\) to \(\beta=0.40\) at \(r=1/16\), falling into the
AdamW regime. Soft-rank scaling degrades more gradually
(\(0.95\rightarrow0.72\)) and remains above AdamW, indicating that low update
rank primarily limits dominant-mode hard capacity rather than all spectral
growth.}
\label{fig:dion_rank_sweep_tail}
\end{figure}

\paragraph{Low update rank primarily limits rare-token hard capacity.}
Figure~\ref{fig:dion_rank_sweep_tail} shows the scaling effect in the TAIL
tokens. As the Dion rank fraction decreases from $r=1/2$ to $r=1/16$, TAIL
hard-rank scaling drops from $\beta_{\mathrm{hard}}=0.88$ to $0.40$, bringing
it closer to the AdamW. By contrast, TAIL soft-rank scaling degrades more
gradually, from $\beta_{\mathrm{soft}}=0.95$ to $0.72$, and remains above AdamW
throughout the sweep. Thus, aggressive rank reduction does not eliminate
diffuse spectral growth, it primarily limits the conversion of added
FFN width into dominant-mode hard capacity.

\paragraph{Hard--soft asymmetry exposes the rank bottleneck.}
The selective degradation of hard rank appears directly in the hard--soft
asymmetry. In the TAIL regime, $\Delta_{1,2}$ increases from $+0.07$ at
$r=1/2$ to $+0.31$ at $r=1/16$,  indicating that added FFN widths are less effectively converted
into dominant-mode capacity. This effect is frequency-dependent: TAIL shows
the clearest asymmetry increase, MID changes more mildly, and HEAD is
non-monotonic across rank fractions. Thus, rank bottleneck is not a
uniform loss of capacity, rather a rare-token hard-capacity ceiling.


\begin{wraptable}{r}{0.48\columnwidth}
\caption{Low update rank increases hard--soft scaling asymmetry (Figure \ref{fig:dion_rank_sweep_tail}).
Aggregate $\Delta_{1,2}$ rises from $+0.13$ at $r{=}1/2$ to $+0.24$ at $r{=}1/16$;
TAIL shows the largest rise ($+0.07{\to}+0.31$).} \vspace{-0.5em}
\label{tab:dion_rank_sweep_asymmetry}
\centering
\footnotesize
\setlength{\tabcolsep}{3pt}
\begin{tabular}{lccccc}
\toprule
 & & \multicolumn{4}{c}{Dion rank $r$} \\
\cmidrule(lr){3-6}
$\Delta_{1,2}$ & AdamW & $\tfrac{1}{2}$ & $\tfrac{1}{4}$ & $\tfrac{1}{8}$ & $\tfrac{1}{16}$ \\
\midrule
Aggregate & $+0.37$ & $+0.13$ & $+0.12$ & $+0.20$ & $+0.24$ \\
\midrule
HEAD & $+0.18$ & $+0.37$ & $+0.17$ & $+0.19$ & $+0.35$ \\
MID  & $+0.21$ & $+0.15$ & $+0.19$ & $+0.20$ & $+0.22$ \\
TAIL & $+0.18$ & $+0.07$ & $+0.18$ & $+0.29$ & $+0.31$ \\
\bottomrule
\end{tabular}
\end{wraptable}

This shows that update rank acts as a control knob
between high-rank, Muon-like capacity scaling and low-rank, AdamW-like
scaling. With sufficient update rank, orthonormalized updates
support growth in both soft and hard spectral capacity. Under aggressive rank
reduction, soft-rank growth remains relatively robust, but rare-token hard-rank
scaling falls toward the AdamW regime. Thus, {\em optimizer-update rank is part of
the representation-scaling design space, not merely an efficiency parameter.}
Full scaling exponents across frequency regimes are reported in Appendix
Table~\ref{tab:dion_rank_sweep_betas}.


\subsection{Optimizer-Dependent Spectral Scaling Persists at Larger Scale}
\label{sec:results:scale}

We next study how the optimizer-dependent scaling laws trends persist at a larger model scale. We 
repeat the core TAIL-token spectral-scaling experiment on GPT-2 350M with a
four-point FFN-width sweep. This coarser width sweep serves the scale-replication check of the optimizer-dependent
ordering rather than a replacement for the more detailed 160M analysis.

\begin{figure}[htbp]
    \centering
    \includegraphics[width=0.49\textwidth]{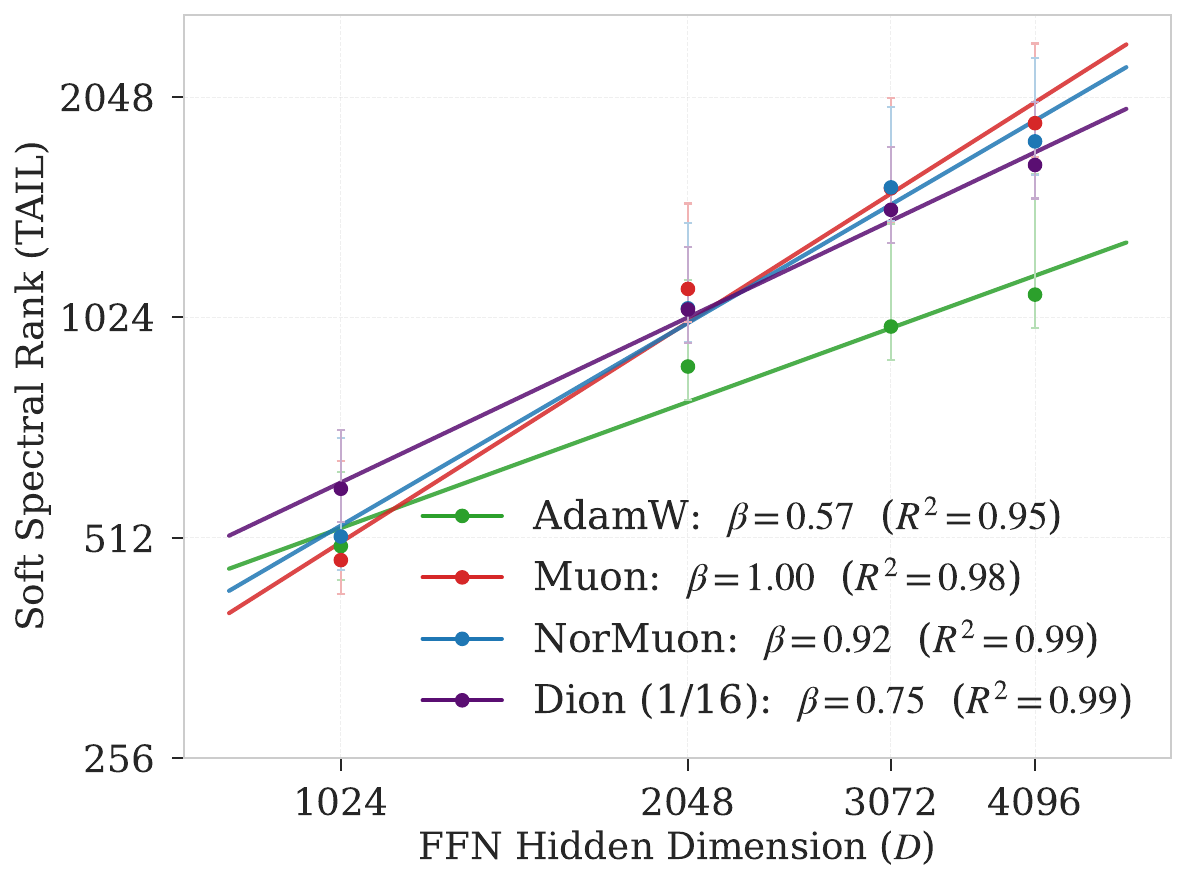}
    \includegraphics[width=0.49\textwidth]{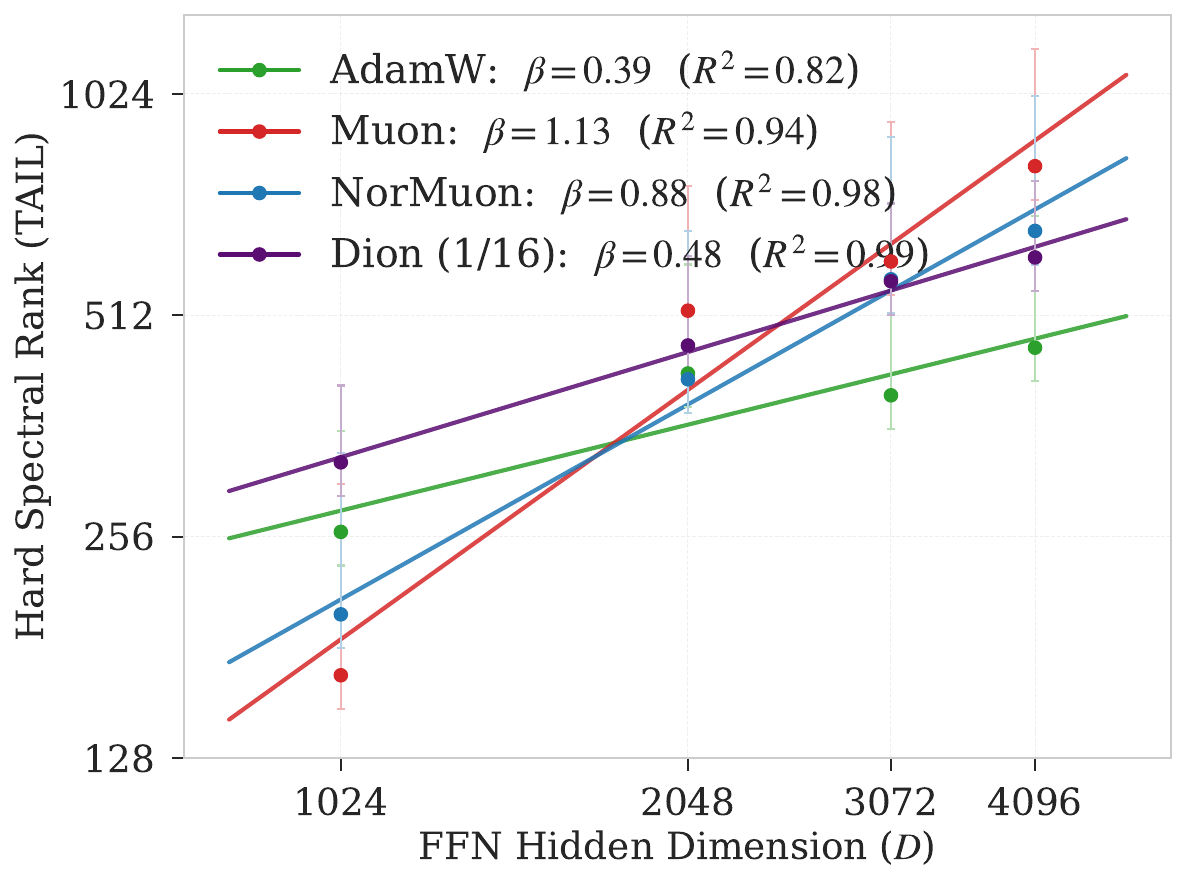}
    \vspace{-0.5em}
    \caption{Optimizer-dependent TAIL spectral scaling persists at 350M
scale. Soft spectral rank (left) and hard spectral rank (right) are shown
across a four-point FFN-width sweep for TAIL tokens in GPT-2 350M. Muon and
NorMuon maintain stronger hard-rank scaling than AdamW, while low-rank
Dion($1/16$) remains in a lower hard-capacity ($\beta_{\mathrm{hard}}$), while their soft-rank scaling incurred less degradation, indicating that
low update rank primarily limits dominant-mode hard capacity.}
    \label{fig:350m_scaling}
\end{figure}

\paragraph{The optimizer-dependent structure persists at 350M.}
Figure~\ref{fig:350m_scaling} shows TAIL-token spectral scaling at 350M, and the
qualitative ordering matches the 160M results. Muon achieves near-linear
hard-rank scaling ($\beta_{\mathrm{hard}}=1.13$, $R^2=0.94$), NorMuon remains
strong ($\beta_{\mathrm{hard}}=0.88$, $R^2=0.98$), and AdamW remains clearly
sublinear ($\beta_{\mathrm{hard}}=0.39$, $R^2=0.82$). Dion~$(1/16)$ is also
sublinear ($\beta_{\mathrm{hard}}=0.48$), far below Muon and NorMuon, showing
that the low-rank update bottleneck persists at larger scale. Thus, the key
optimizer-dependent scaling trends observed at 160M, strong $\beta_{\mathrm{hard}}$
for Muon/NorMuon and weaker for AdamW/low-rank Dion, also
appear at 350M.

\begin{wraptable}{r}{0.65\columnwidth}
\vspace{-\intextsep}
\caption{Scale consistency from 160M to 350M: each cell reports the 160M/350M value. Muon and NorMuon maintain stronger
hard-rank scaling than AdamW and low-rank Dion across scale, while AdamW and
Dion($1/16$) retain positive  asymmetry ($\Delta_{1,2}$).} \vspace{-0.6em}
\label{tab:scale_comparison}
\centering
\small
\setlength{\tabcolsep}{3pt}
\begin{tabular}{l cc cc}
\toprule
 & \multicolumn{2}{c}{Aggregate} & \multicolumn{2}{c}{TAIL} \\
\cmidrule(lr){2-3} \cmidrule(lr){4-5}
Optimizer    & $\beta_{\mathrm{hard}}$ & $\Delta_{1,2}$ & $\beta_{\mathrm{hard}}$ & $\Delta_{1,2}$ \\
\midrule
AdamW        & 0.29 / 0.46 & $+0.37$ / $+0.17$ & 0.44 / 0.39 & $+0.18$ / $+0.19$ \\
Muon         & 0.82 / 1.21 & $+0.14$ / $-0.16$ & 1.02 / 1.13 & $+0.01$ / $-0.13$ \\
NorMuon      & 0.80 / 0.95 & $+0.21$ / $+0.02$ & 1.04 / 0.88 & $+0.00$ / $+0.04$ \\
Dion (1/16)  & 0.50 / 0.53 & $+0.24$ / $+0.30$ & 0.40 / 0.48 & $+0.31$ / $+0.27$ \\
\bottomrule
\end{tabular}
\end{wraptable}

\paragraph{Hard--soft asymmetry also generalizes.}
Table~\ref{tab:scale_comparison} compares $\beta_{\mathrm{hard}}$ and
$\Delta_{1,2}$ between 160M and 350M models. AdamW maintains positive TAIL
asymmetry across scale ($+0.18/+0.19$), indicating that added FFN width converts 
primarily into the diffuse capacity scaling rather than dominant-mode capacity scaling. 
Dion~$(1/16)$ also retains elevated positive asymmetry ($+0.31/+0.27$),
consistent with the rank-bottleneck behavior in
Section~\ref{subsec:dion_rank_sweep}. By contrast, Muon reaches slightly
negative asymmetry at 350M in both aggregate and TAIL spectra
($\Delta_{1,2}=-0.16$ and $-0.13$), while NorMuon remains close to zero
asymmetry. Overall, the 350M experiment supports the conclusion that optimizer
geometry shapes representation scaling beyond a single model size.


\subsection{Optimizer-Induced Scaling Effects Dominate the Attention-Rank Interventions}
\label{sec:results:optimizer_vs_architecture}

Having established that optimizer choice changes how FFN width is converted into realized
capacity for a fixed architecture, we next ask whether these optimizer-induced
effects are comparable to, or larger than, a controlled architectural
intervention. Motivated by recent work showing that attention rank can limit
expressivity~\citep{amsel2025quality,garg2022can}, we increase per-head
attention rank by reducing the number of attention heads, which increases the attention ranks
while preserving the total parameter count.

\begin{figure}[t]
\centering
\includegraphics[width=\linewidth]{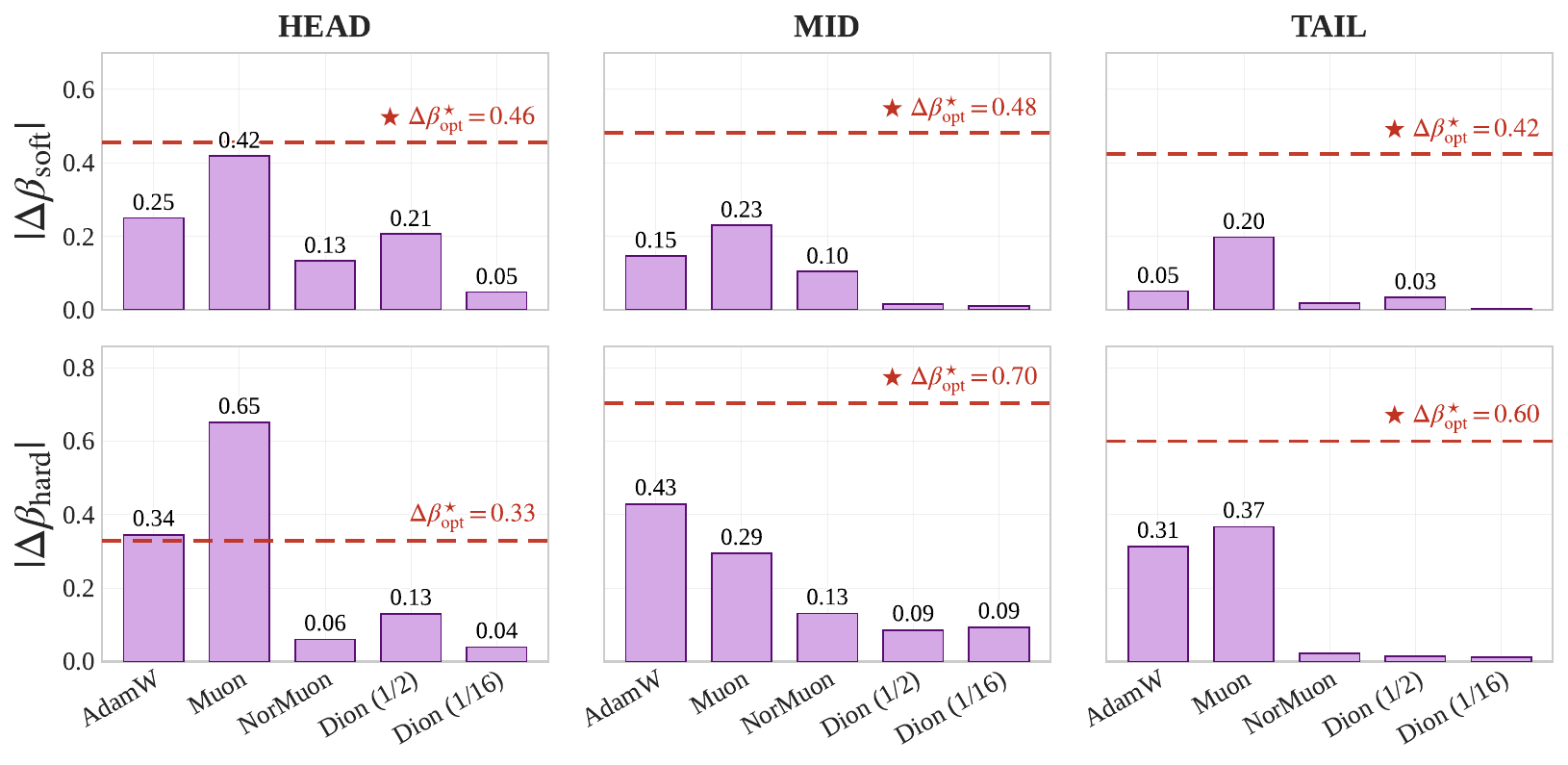}
\vspace{-2em}
\caption{Optimizer-induced shifts in spectral-scaling exceed attention-rank shifts in GPT-2 160M.
We compare the optimizer-induced spectral scaling shift for AdamW in the original $12$-head 
architecture, $\Delta\beta_{\mathrm{opt}}^{\star}$ (shown as red dashed lines), and the spectral scaling shift induced by architectural 
intervention, increasing the attention ranks at fixed total parameter
count, across each optimizer (shown as bars). Optimizer-induced gains exceed attention-rank shifts in $5$ of $6$ frequency regimes (marked with \textcolor{red}{$\bigstar$}), across $28$ of $30$ scaling shifts; the only exceptions occur in HEAD
hard-rank scaling.}
\label{fig:optimizer_vs_architecture}
\end{figure}

For each frequency regime $b$ and spectral rank metric $m$, we compare two effects on
the fitted scaling exponent $\beta$. The optimizer-induced gain over AdamW in the original $12$-head
architecture as: \vspace{-0.6em}
\begin{equation}
    \Delta\beta_{\mathrm{opt}}^{\star,(b,m)}
    =
    \max_o \beta_{o,12h}^{(b,m)}
    -
    \beta_{\mathrm{AdamW},12h}^{(b,m)}.
\end{equation} 
where the maximum is taken over optimizers. The attention-rank induced spectral scaling shift for optimizer $o$ is \vspace{-1em}
\begin{equation}
    A_{\mathrm{rank}}^{(b,m)}(o)
    =
    \left|
    \beta_{o,6h}^{(b,m)}
    -
    \beta_{o,12h}^{(b,m)}
    \right|.
\end{equation} 

Further, to test how uniformly architectural intervention is expressed across optimizers, we report the signed
architectural effect for each optimizer $o$, frequency regime $b$, and rank metric $m$ as:  \vspace{-0.6em}
\begin{equation} \label{eqn:signed_arch_effects}
    \Delta\beta_{\mathrm{arch}}^{(b,m)}(o)
    =
    \beta_{o,6h}^{(b,m)}
    -
    \beta_{o,12h}^{(b,m)} .
\end{equation}
These comparisons investigate whether optimizer choice changes the spectral-scaling exponents 
more than the tested architectural intervention, such as increasing attention-rank under fixed parameter count, 
and whether the intervention is expressed uniformly across optimizers.

\paragraph{Optimizer effects exceed attention-rank effects in nearly all regimes.}
Figure~\ref{fig:optimizer_vs_architecture} shows that the optimizer-induced gain
is larger than the attention-rank shift in $28$ of $30$ 
comparisons. The only exceptions occur in
$\beta_{\mathrm{hard}}(\mathrm{HEAD})$, where the attention-rank shifts under
AdamW ($0.345$) and Muon ($0.651$) exceed the optimizer-induced gain
($0.330$). Thus, {\em HEAD hard-rank scaling is the most architecture-sensitive
regime under this intervention.} However, in MID and TAIL regimes the
optimizer-induced shift exceeds the attention-rank-induced shifts for every optimizer.
For instance, the optimizer-induced $\beta_{\mathrm{hard}}(\mathrm{MID})$ gain is $0.703$, 
exceeding the largest attention-rank shift ($0.43$ under AdamW); and  
$\beta_{\mathrm{hard}}(\mathrm{TAIL})$ gain is  $0.6$, exceeding the largest shift of $0.367$ under Muon.

\begin{wraptable}{r}{0.55\columnwidth}
\centering
\footnotesize
\setlength{\tabcolsep}{1.25pt}
\caption{Signed attention-rank architectural effects (Eq. \ref{eqn:signed_arch_effects})
in GPT-2 160M. Both, the sign and magnitude of the architectural effect are strongly optimizer-dependent.} \vspace{-1em}
\label{tab:signed_head_delta}
\begin{tabular}{lccccc}
\toprule
Regime & AdamW & Muon & NorMuon & Dion(1/2) & Dion(1/16) \\
\midrule
$\beta_{\mathrm{hard}}(\mathrm{HEAD})$ & +0.345 & +0.651 & +0.060 & -0.130 & +0.039 \\
$\beta_{\mathrm{soft}}(\mathrm{HEAD})$ & +0.250 & +0.419 & -0.133 & -0.206 & -0.048 \\
$\beta_{\mathrm{hard}}(\mathrm{MID})$  & +0.429 & +0.294 & +0.131 & -0.086 & -0.093 \\
$\beta_{\mathrm{soft}}(\mathrm{MID})$  & +0.147 & +0.230 & +0.104 & +0.015 & -0.011 \\
$\beta_{\mathrm{hard}}(\mathrm{TAIL})$ & +0.313 & +0.367 & -0.023 & -0.014 & -0.011 \\
$\beta_{\mathrm{soft}}(\mathrm{TAIL})$ & +0.051 & +0.198 & +0.019 & -0.034 & -0.002 \\
\bottomrule
\end{tabular}
\end{wraptable}
\paragraph{Attention-rank-induced scaling shifts depend on optimizer geometry.}
The signed effect (Eq. \ref{eqn:signed_arch_effects}) shows that this architectural intervention is not expressed
uniformly across optimizers. For $\beta_{\mathrm{hard}}(\mathrm{TAIL})$,
increasing per-head attention rank raises the exponent under AdamW and Muon by
$+0.313$ and $+0.367$, respectively, but has near-zero effect under NorMuon and
Dion variants ($|\Delta\beta_{\mathrm{arch}}|\leq 0.023$). Moreover, this effect is also
selective across spectral ranks. For TAIL tokens, attention-rank shift is
larger for $\beta_{\mathrm{hard}}$ than for $\beta_{\mathrm{soft}}$ under AdamW
($+0.313$ vs. $+0.051$) and Muon ($+0.367$ vs. $+0.198$). This suggest that
increased per-head rank {\em primarily affects dominant-mode capacity} rather than
diffuse spectral spread.

This intervention also changes the optimizer--architecture match. In the original 
12-head architecture, NorMuon attains the largest scaling exponent in five of six
token-regimes; however,  in 6-head architecture, Muon attains the largest exponent in
all six regimes. Thus, increasing per-head attention rank does not impose a
fixed spectral shift across optimizers. Rather, the architectural intervention
changes which optimizer most effectively converts added FFN width into spectral
capacity. Refer to Appendix~\ref{app:optimizer_architecture_effects} for more in-depth  analysis.


\subsection{Removing RoPE Reshapes Optimizer-Dependent Spectral Scaling}
\label{sec:results:nope}

We next study a second architectural intervention, removing explicit positional
signal. RoPE injects relative positional structure into self-attention
\citep{su2024roformer}, while recent work shows that Transformers can operate
without explicit positional embeddings and may benefit from NoPE designs
\citep{kazemnejad2023impact,chi2023latent,gelberg2026extending}. We evaluate
this intervention at 350M scale for Muon and NorMuon, whose 
power-law fits, for both RoPE and NoPE, are reliable across frequency regimes. 
More importantly, this comparison separates two related optimizer geometries: both optimizers use orthonormalized
updates, while NorMuon additionally imposes per-neuron normalization.


\begin{wraptable}{r}{0.55\columnwidth}
\vspace{-\intextsep}
\caption{Removing RoPE induces optimizer-dependent spectral
redistribution. Hard-rank scaling exponents and hard--soft asymmetry are reported for GPT-2
350M RoPE/NoPE models. Removing RoPE under Muon increases
$\beta_{\mathrm{hard}}$ for HEAD, but decreases for MID and TAIL tokens. Under NorMuon  
it decreases across all token regimes. All reported fits have $R^2\geq0.93$.} \vspace{-0.6em}
\label{tab:rope_nope_350m}
\centering
\small
\setlength{\tabcolsep}{2pt}
\begin{tabular}{ll ccc}
\toprule
Optimizer &  & $\beta_{\mathrm{hard}}$ & $\Delta\beta_{\mathrm{hard}}$ & $\Delta_{1,2}$ \\
\midrule
\multirow{3}{*}{Muon}
 & HEAD & 1.083 / 1.286 & $+0.203$ & $-0.201$ / $-0.004$ \\
 & MID  & 1.008 / 0.744 & $-0.264$ & $-0.085$ / $+0.096$ \\
 & TAIL & 1.127 / 0.836 & $-0.291$ & $-0.129$ / $+0.091$ \\
\midrule
\multirow{3}{*}{NorMuon}
 & HEAD & 1.023 / 0.656 & $-0.367$ & $-0.045$ / $+0.315$ \\
 & MID  & 0.723 / 0.569 & $-0.154$ & $+0.145$ / $+0.276$ \\
 & TAIL & 0.881 / 0.739 & $-0.142$ & $+0.038$ / $+0.136$ \\
\bottomrule
\end{tabular}
\end{wraptable}

\paragraph{NoPE produces optimizer-dependent capacity redistribution.}
Table~\ref{tab:rope_nope_350m} shows that removing RoPE does not induce a
uniform change in spectral scaling. The clearest sign reversal occurs in HEAD
hard-rank scaling, NoPE increases Muon's exponent from $1.083$ to $1.286$
($\Delta\beta=+0.203$), but decreases NorMuon's from $1.023$ to $0.656$
($\Delta\beta=-0.367$). Under Muon, NoPE shifts hard-capacity scaling toward
HEAD tokens while reducing MID and TAIL scaling
($1.008\rightarrow0.744$ and $1.127\rightarrow0.836$). Under NorMuon, NoPE
decreases hard-rank scaling in all three frequency regimes, with the largest
drop in HEAD. Thus, hard--soft asymmetry also changes: under Muon, MID
and TAIL asymmetry shift from negative to positive
($-0.085\rightarrow+0.096$ and $-0.129\rightarrow+0.091$), while under NorMuon
asymmetry increases in all three regimes, with the largest increase in HEAD
($+0.045\rightarrow+0.315$). 

In other words, positional-signal removal changes not only
the magnitude of hard-rank scaling, but also how added FFN width is converted into
dominant-mode vs diffuse spectral capacity. Appendix~\ref{app:positional} provides a position-dependent FFN analysis
supporting this interpretation.

\paragraph{NoPE redistributes optimizer gaps across token-frequency regimes.}
Removing RoPE also changes where optimizer differences appear, across the token-frequency regimes. The Muon--NorMuon
hard-rank gap widens sharply in HEAD tokens, from
$|1.083-1.023|=0.06$ under RoPE to $|1.286-0.656|=0.63$ under NoPE, because
NoPE improves Muon while degrading NorMuon. In contrast, the gap narrows in MID
($|1.008-0.723|=0.285$ to $|0.744-0.569|=0.175$) and TAIL
($|1.127-0.881|=0.246$ to $|0.836-0.739|=0.097$). Thus, removing positional
signal redistributes optimizer-dependent spectral capacity across the token-frequency
spectrum.


\begin{wraptable}{r}{0.5\columnwidth}
\vspace{-\intextsep}
\centering
\small
\setlength{\tabcolsep}{3pt}
\caption{Effect-size synthesis for TAIL hard-rank scaling.
We compare the magnitude of the AdamW$\rightarrow$Muon optimizer gap with
TAIL hard-rank shifts induced by attention-rank (160M) and RoPE/NoPE interventions (350M).
} \vspace{-0.8em}
\label{tab:effect_summary_tail_hard}
\begin{tabular}{lcc}
\toprule
Intervention & $|\Delta\beta_{\mathrm{hard}}|$ & Relative size \\
\midrule
AdamW $\rightarrow$ Muon & 0.74 & $1.00\times$ \\
$12h \rightarrow 6h$ (Muon) & 0.37 & $0.50\times$ \\
RoPE $\rightarrow$ NoPE (Muon) & 0.29 & $0.39\times$ \\
RoPE $\rightarrow$ NoPE (NorMuon) & 0.14 & $0.19\times$ \\
\bottomrule
\end{tabular}
\end{wraptable}

\paragraph{Optimizer choice remains the larger TAIL hard-rank effect.}
Table~\ref{tab:effect_summary_tail_hard} compares effect sizes in the TAIL
hard-rank regime. The AdamW$\rightarrow$Muon optimizer shift
($|\Delta\beta|=0.74$) is about $2.0\times$ the attention-rank effect and
$2.5$--$5.2\times$ the RoPE$\rightarrow$NoPE effects. Thus, optimizer choice
remains the dominant source of variation in the TAIL hard-rank comparison among
the measured interventions. At the same time, the RoPE/NoPE results show that
architectural signal removal reshapes the optimizer gap itself. 

These results show that the effect of architectural interventions on representational scaling is mediated by optimizer geometry. 
Increasing per-head attention rank can alter optimizer ordering, while removing RoPE
redistributes optimizer gaps across token-frequency regimes. Thus, both
capacity-axis and positional-signal interventions reshape spectral scaling
through the architecture--optimizer pair rather than acting as
independent perturbations.

\section{Discussion and Conclusion}
Spectral scaling laws are not properties of architecture alone, optimizer
geometry changes how added FFN width is expressed in representation space. The
main distinction between optimizers is not whether spectral rank grows, but
which kind of spectral capacity grows. Under AdamW, added FFN width contributes
primarily to diffuse spectral capacity rather than dominant-mode capacity. 
Muon-style optimizers make this conversion
substantially more efficient, with the largest gains concentrated in MID- and
TAIL-frequency token regimes. Dion rank sweeps further show that update rank
controls this conversion, orthonormalization alone is insufficient when the
update rank is aggressively constrained, and low-rank Dion approaches
AdamW-like scaling. These effects persist beyond convergence
differences, matched validation loss can coexist with distinct 
scaling behavior, showing that optimizer-induced representation geometry is not
a transient training artifact.

The broader implication is that effective representational capacity is realized
by an architecture--optimizer pair, not by architecture alone. Architectural
interventions such as increasing per-head attention rank and removing RoPE do
not induce fixed, optimizer-independent spectral shifts. Their effects are
mediated by optimizer geometry; in some regimes, optimizer-induced gains exceed
those of architectural interventions, while in others, architectural changes
redistribute where optimizer differences appear across token-frequency regimes
or change which optimizer is best matched to a given architecture. Optimization
should therefore be treated as part of the model-design space, not as a
training procedure applied after the architecture has been chosen. Future
scaling-law analyses should treat optimizer geometry as a first-class axis
alongside model size, data, compute, and architecture.

{\bf Limitations.}
Our study establishes spectral scaling laws within a controlled empirical
regime, using GPT-style decoder-only models with 160M and 350M parameters.
Although the 350M runs replicate the main 160M patterns, testing at 1B+ scale
would further strengthen the evidence for frontier-scale generality. This is
compute-intensive, a single scaling-law measurement requires training a sweep of
FFN-width variants, up to eight models per optimizer and model scale in our
main setting. Our architecture--optimizer co-design analysis also focuses on
dense FFN architectures trained with AdamW, Muon, NorMuon, and Dion variants
under a fixed data recipe. Other architectures, training protocols, and
optimizer families may produce different spectral scaling behavior.

Further, soft and hard spectral ranks,
along with the R\'enyi-family effective-capacity measures in
Appendix~\ref{sec:renyi_family}, quantify how FFN width is converted into
usable capacity, but they do not fully characterize downstream task behavior. 
Within our scaling-law analyses, hard-rank scaling
and TAIL-token regimes emerge as especially sensitive views of
optimizer-dependent capacity allocation. Testing whether
targeted interventions on update geometry, or spectral
concentration can causally control representation capacity, remain important
directions for future work.

\newpage
\bibliographystyle{unsrt}
\bibliography{MyRef}



\newpage
\appendix
\section*{Appendix}
\startcontents[appendix]
\printcontents[appendix]{}{1}{\setcounter{tocdepth}{2}}

\newpage

\section{Experimental Setup}
\label{app:experimental_setup}

\subsection{Model architecture}
\label{app:arch}

All models follow the modded-nanoGPT~\citep{modded_nanogpt_2024}
configuration, summarized in Table~\ref{tab:arch}. All linear layers
(attention projections, FFN, and the LM head) are bias-free. Attention uses
rotary position embeddings~\citep{su2024roformer} with base $10{,}000$,
applied per-head, and a non-parametric RMSNorm is additionally applied to
the query and key tensors after RoPE (QK-norm). The FFN block uses two
linear projections with squared-ReLU~\citep{so2021searching} between them.
Pre-Norm and Post-Norm placements are controlled by a single
\texttt{postln\_frac} parameter $k \in [0, 1]$: the first
$\lfloor k \cdot L \rfloor$ layers use Post-RMSNorm and the remainder use
Pre-RMSNorm, with $k = 0$ recovering pure Pre-RMSNorm; this parameterization
is used in the partial PostLN  configurations. Weights are initialized with the
spectral-condition scheme of~\cite{yang2023spectral},
$\sigma = (1/\sqrt{d_{\text{in}}})\min(1, \sqrt{d_{\text{out}}/d_{\text{in}}})$.

\begin{table}[h]
\centering
\caption{Architectural configurations. All models use the GPT-2 byte-pair
tokenizer with padded vocabulary size 50{,}304. The 160M and 350M labels
denote the base $4\times$ FFN configuration; FFN-width sweeps vary only the
FFN intermediate dimension.}
\label{tab:arch}
\small
\begin{tabular}{lcccc}
\toprule
Model & Layers & Heads & $d_{\text{model}}$ & Base FFN width ($D$) \\
\midrule
160M & 12 & 12 & 768  & 3072 \\
350M & 24 & 32 & 1024 & 4096 \\
\bottomrule
\end{tabular}
\end{table}

\subsection{Training protocol}
\label{app:training}

Table~\ref{tab:training} summarizes the training details. All models are
trained with a constant learning rate followed by a 20\% linear cooldown to
zero, no warmup, BF16 mixed precision, and no
gradient clipping. Each (optimizer, FFN width) combination is trained as an
independent run. For the extended-training ~\ref{sec:results:matched_loss}, AdamW
models are trained for 12{,}000 iterations with all other settings held
fixed. Validation perplexity is computed on a
held-out FineWeb-Edu subset of $10.5$M tokens.

\begin{table}[h]
\centering
\caption{Training protocol. Tokens per step $= 1024 \times 512 = 524{,}288$.
The 160M and 350M labels denote the base $4\times$ FFN configuration; FFN
width is varied as $D = m\, d_{\text{model}}$ with $m$ as listed below, so
total parameter count and per-iteration cost scale with $D$ while all other
settings are held fixed.}
\label{tab:training}
\small
\begin{tabular}{lcc}
\toprule
& 160M & 350M \\
\midrule
Dataset                       & FineWeb-Edu~\citep{penedo2024fineweb} & FineWeb-Edu~\citep{penedo2024fineweb} \\
Sequence length               & 512          & 512          \\
Global batch size             & 1024         & 1024         \\
Iterations                    & 6{,}000      & 8{,}000      \\
Total tokens                  & 3.15B        & 4.19B        \\
\bottomrule
\end{tabular}
\end{table}

\subsection{Optimizer hyperparameters}
\label{app:opt_hp}

All matrix-aware optimizers (Muon~\citep{jordan2024muon},
NorMuon~\citep{li2025normuon}, Dion~\citep{ahn2025dion}) follow the
parameter-update recipe described in~\citep{ahn2025dion}:
Lion~\citep{chen2023symbolic} is used as the scalar optimizer for
non-matrix parameters (embeddings, normalization scales, the LM head), and
the per-parameter learning-rate scaling follows the spectral-condition
prescription of~\citep{yang2023spectral}. Optimizer hyperparameters are listed in
Table~\ref{tab:opt_hp}. Learning-rate ablations
(Appendix~\ref{app:lr_ablation}) additionally sweep AdamW learning rate over
$\{1, 3, 6\} \times 10^{-3}$ and Muon learning rate over
$\{1, 2, 4\} \times 10^{-2}$, holding all other hyperparameters fixed.

\begin{table}[h]
\centering
\caption{Optimizer hyperparameters. Weight decay is $0.01$ throughout;
momentum $\mu = 0.95$ for Muon, NorMuon, and Dion. AdamW
\citep{loshchilov2018decoupled} uses $(\beta_1, \beta_2) = (0.9, 0.95)$.}
\label{tab:opt_hp}
\small
\begin{tabular}{lcccc}
\toprule
                     & AdamW              & Muon               & NorMuon            & Dion \\
\midrule
Learning rate        & $3{\times}10^{-3}$ & $2{\times}10^{-2}$ & $2{\times}10^{-2}$ & $2{\times}10^{-2}$ \\
LR scaling           & ---                & spectral-norm      & spectral-norm      & spectral-norm \\
Scalar optimizer     & ---                & Lion~\citep{chen2023symbolic} & Lion~\citep{chen2023symbolic} & Lion~\citep{chen2023symbolic} \\
\bottomrule
\end{tabular}
\end{table}

\subsection{Token-frequency stratification}
\label{app:freq_stratification}

Token frequencies $f(v)$ are computed once from the FineWeb-Edu tokenized
training shards (10.26B tokens total) by histogram aggregation over GPT-2
byte-pair token IDs and held fixed across all experiments, ensuring that
HEAD/MID/TAIL regime assignments are identical across optimizers, FFN
widths, and model scales. Tertile thresholds $\tau_{\text{head}}$ and
$\tau_{\text{mid}}$ are set by cumulative occurrence mass: sorting token
types by decreasing $f(v)$ and taking cumulative sums, $\tau_{\text{head}}$
is the smallest frequency such that cumulative mass remains below $M/3$,
and $\tau_{\text{mid}}$ is defined analogously at $2M/3$
(Eq.~\ref{eq:freq_stratification}). The resulting regimes each cover
approximately one third of total occurrences but span dramatically
different numbers of token \emph{types}, reflecting the heavy-tailed
Zipfian structure of natural language~\citep{kandpal2023large,
kunstner2025scaling}: 30 token types (0.06\% of vocabulary) carry the HEAD
third of occurrences, 1{,}215 types (2.42\%) carry the MID third, and
49{,}059 types (97.53\%) carry the TAIL third. regime statistics are
summarized in Table~\ref{tab:freq_regimes}. 


\begin{table}[t]
\centering
\caption{Frequency-regime statistics for FineWeb-Edu (GPT-2 BPE, vocab
50{,}304; corpus size $\approx 10.26$B tokens). Tertiles are set by
cumulative occurrence mass.}
\label{tab:freq_regimes}
\small
\begin{tabular}{lcccc}
\toprule
regime & Frequency threshold & Token types & \% of vocabulary & \% of occurrences \\
\midrule
HEAD & $f(v) \geq 32{,}815{,}898$                            &       30 &  0.06\% & 32.79\% \\
MID  & $900{,}178 \leq f(v) < 32{,}815{,}898$                & 1{,}215  &  2.42\% & 34.21\% \\
TAIL & $f(v) < 900{,}178$                                    & 49{,}059 & 97.53\% & 33.00\% \\
\bottomrule
\end{tabular}
\end{table}

\subsection{Spectral measurements and reporting}
\label{app:spectral_measurement}

Pre- and post-activation FFN representations
(Section~\ref{subsec:probe_points}) are collected on the held-out validation
batch ($\approx$10.5M tokens), aggregated across batch and sequence positions
before computing the empirical covariance and its normalized eigenspectrum
(Eq.~\ref{eq:covariance_eigenspectrum}). Spectral-rank quantities are
computed per layer and then averaged across layers; frequency-stratified
results are computed separately within each HEAD/MID/TAIL regime before
layer averaging. Eigen-metric statistics are logged every 200 training
steps at the 160M scale and every 400 steps at the 350M scale. For each scaling fit,
we report the fitted exponent $\beta$ and the corresponding coefficient of
determination $R^2$. All scaling-law fits use single seeds per (optimizer,
FFN width) configuration; error bars in figures correspond to the
inter-layer standard deviation of per-layer measurements.

\subsection{Compute}
\label{app:compute}

We use 4$\times$NVIDIA
RTX 3090 GPUs (24~GB each) for the 160M-scale experiments and
8$\times$NVIDIA RTX 3090 GPUs for the 350M-scale experiments. The main FFN-width
sweep comprises 40 training runs at 160M (five optimizers $\times$ eight
widths) and 20 runs at 350M (five optimizers $\times$ four widths),
complemented by the learning-rate, normalization-placement,
positional-encoding, and extended-training ablations.

\section{Validation Perplexity Across Spectral-Scaling Runs}
\label{app:eval_ppl_full}

In this section, we report the validation perplexities (PPL) for the training runs used in the
spectral-scaling analyses. All values are computed on the held-out
validation split of FineWeb-Edu~\citep{penedo2024fineweb} with context length 512, following the evaluation
protocol in Appendix~\ref{app:experimental_setup}. These values provide
loss-level context for the representation-geometry results. Note that the main
comparisons are not between failed and successful training runs, rather between
runs that can achieve comparable validation perplexity while exhibiting
different spectral-scaling behavior. 

\subsection{GPT-2 160M FFN-Width Sweep}
\label{app:ppl_160m}

Table~\ref{tab:appendix_eval_ppl_160m} reports validation perplexity across the
full eight-point FFN-width sweep for the 160M base configuration. The AdamW
12K row is included as the extended-training experiments described in
Section~\ref{sec:results:matched_loss}. AdamW improves substantially with longer
training and reaches perplexity comparable to Dion~($r=1/16$), but remains
well behind Muon and NorMuon optimizers.

\begin{table}[h]
\centering
\caption{Validation perplexity for the GPT-2 160M FFN-width sweep. Perplexity
is reported across FFN widths on the held-out validation set. AdamW 12K is
included as an extended-training control.}
\label{tab:appendix_eval_ppl_160m}
\begin{tabular}{l cccccccc}
\toprule
 & \multicolumn{8}{c}{FFN width ($d = 768$)} \\
\cmidrule(lr){2-9}
Optimizer          & $d$   & $2d$  & $3d$  & $4d$  & $5d$  & $6d$  & $7d$  & $8d$  \\
\midrule
AdamW (6K)         & 38.15 & 36.79 & 34.12 & 34.34 & 33.85 & 31.68 & 32.17 & 32.43 \\
AdamW (12K)        & 34.25 & 32.15 & 30.40 & 30.47 & 29.43 & 29.04 & 28.45 & 28.29 \\
Muon               & 31.79 & 29.83 & 28.65 & 27.90 & 27.27 & 26.85 & 26.43 & 26.15 \\
NorMuon            & 31.77 & 29.82 & 28.63 & 27.82 & 27.23 & 26.74 & 26.29 & 25.94 \\
\midrule
Dion ($r{=}1/2$)   & 32.06 & 30.19 & 28.96 & 28.28 & 27.67 & 27.16 & 26.83 & 26.51 \\
Dion ($r{=}1/4$)   & 32.61 & 30.69 & 29.58 & 28.79 & 28.23 & 27.68 & 27.41 & 27.13 \\
Dion ($r{=}1/8$)   & 33.21 & 31.34 & 30.19 & 29.51 & 28.94 & 28.53 & 28.22 & 27.89 \\
Dion ($r{=}1/16$)  & 34.18 & 32.41 & 31.33 & 30.66 & 30.08 & 29.71 & 29.40 & 29.23 \\
\bottomrule
\end{tabular}
\end{table}

\subsection{Architectural Intervention: Increasing Attention Ranks in GPT-2 160M}
\label{app:ppl_160m_6h}

Table~\ref{tab:appendix_eval_ppl_160m_6h} reports the validation
perplexities for the 6-head architecture used in
Section~\ref{sec:results:optimizer_vs_architecture}. This intervention changes
the attention-head configuration while preserving the same FFN-width sweep, at fixed parameter count.
Muon and NorMuon remain the top-performing optimizers, while the Dion variants
and AdamW preserve the same broad ordering as in the original 12-head setting.

\begin{table}[h]
\centering
\caption{Validation perplexity for the GPT-2 160M 6-head sweep. Perplexity is
reported across FFN widths for the reduced-head-count architecture.}
\label{tab:appendix_eval_ppl_160m_6h}
\begin{tabular}{l cccccccc}
\toprule
 & \multicolumn{8}{c}{FFN width ($d = 768$)} \\
\cmidrule(lr){2-9}
Optimizer    & $d$   & $2d$  & $3d$  & $4d$  & $5d$  & $6d$  & $7d$  & $8d$  \\
\midrule
AdamW              & 37.95 & 35.40 & 34.36 & 33.34 & 32.95 & 32.65 & 32.49 & 32.19 \\
Muon               & 31.56 & 29.69 & 28.59 & 27.84 & 27.22 & 26.71 & 26.36 & 26.09 \\
NorMuon            & 31.59 & 29.62 & 28.55 & 27.67 & 27.08 & 26.69 & 26.21 & 25.89 \\
\midrule
Dion ($r{=}1/2$)   & 31.93 & 29.96 & 28.99 & 28.26 & 27.59 & 27.20 & 26.77 & 26.47 \\
Dion ($r{=}1/16$)  & 33.91 & 32.09 & 31.16 & 30.50 & 29.94 & 29.50 & 29.27 & 28.94 \\
\bottomrule
\end{tabular}
\end{table}

\subsection{Spectral Scaling at Scale: GPT-2 350M FFN-Width Sweep}
\label{app:ppl_350m}

Table~\ref{tab:appendix_eval_ppl_350m} reports validation perplexity for the
350M scale-replication runs in Section~\ref{sec:results:scale}. The sweep is
coarser than the 160M sweep, but the loss-level ordering remains consistent:
Muon and NorMuon achieve best perplexity, followed by Dion~($r=1/16$), with AdamW
remaining higher in perplexity.

\begin{table}[h]
\centering
\caption{Validation perplexity for the GPT-2 350M FFN-width sweep. Perplexity
is reported across the four FFN widths used in the 350M scale-replication
experiment.}
\label{tab:appendix_eval_ppl_350m}
\begin{tabular}{l cccc}
\toprule
 & \multicolumn{4}{c}{FFN width ($d = 1024$)} \\
\cmidrule(lr){2-5}
Optimizer          & $d$   & $2d$  & $3d$  & $4d$  \\
\midrule
AdamW              & 31.27 & 29.00 & 28.38 & 28.35 \\
Muon               & 26.07 & 24.41 & 23.51 & 22.97 \\
NorMuon            & 26.06 & 24.37 & 23.44 & 22.80 \\
\midrule
Dion ($r{=}1/16$)  & 27.95 & 26.33 & 25.45 & 24.90 \\
\bottomrule
\end{tabular}
\end{table}

\subsection{Architectural Intervention: Removing RoPE in GPT-2 350M}
\label{app:ppl_350m_nope}

Table~\ref{tab:appendix_eval_ppl_350m_nope} reports validation perplexity for
the NoPE runs used in Section~\ref{sec:results:nope}. Removing RoPE increases
perplexity for all optimizers, but Muon and NorMuon remain at the top. 
These values provide the loss-level context for the RoPE vs NoPE architectural intervention.

\begin{table}[h]
\centering
\caption{Validation perplexity for GPT-2 350M without RoPE (Section~\ref{sec:results:nope}).}
\label{tab:appendix_eval_ppl_350m_nope}
\begin{tabular}{l cccc}
\toprule
 & \multicolumn{4}{c}{FFN width ($d = 1024$)} \\
\cmidrule(lr){2-5}
Optimizer    & $d$   & $2d$  & $3d$  & $4d$  \\
\midrule
AdamW        & 33.24 & 30.90 & 29.27 & 29.66 \\
Muon         & 27.05 & 25.33 & 24.37 & 23.75 \\
NorMuon      & 27.08 & 25.26 & 24.29 & 23.60 \\
\bottomrule
\end{tabular}
\end{table}

\subsection{Summary}
\label{app:ppl_summary}

Across the validation-perplexity tables, Muon and NorMuon consistently form the
strongest loss-level optimizer family, while AdamW remains substantially higher
in perplexity. The extended AdamW control improves validation perplexity and
matches Dion~($r=1/16$) closely, however their spectral
geometry remains distinct. Thus, the spectral-scaling differences reported in
the paper are not reducible to a simple failed-optimization explanation.
Architectural interventions such as increasing attention ranks or removing RoPE shift
absolute perplexity, but do not remove the need to analyze optimizer-dependent
representation geometry.

\section{R\'enyi Effective Rank Analysis: Where Optimizer-Induced Capacity Forms}
\label{sec:renyi_family}

The methodology section \ref{subsec:renyi_rank} defines the R\'enyi effective-rank family
$R_\alpha$ as a continuum of spectral-capacity probes. From
information-theoretic viewpoint, the order $\alpha$ changes the resolution at which
the normalized eigenspectrum is probed, lower orders give greater relative
weight to weak eigendirections (see Table \ref{tab:renyi_alpha_interpretation}), while higher orders increasingly emphasize
dominant modes~\citep{renyi1961measures,roy2007effective,van2014renyi,principe2010information}.
We uses two anchors of this family, $R_1$, the soft-rank measure of
diffuse spectral capacity, and $R_2$, the hard-rank measure of
dominant-mode capacity. Now, we perform full R\'enyi-order sweep to study where
optimizer-induced spectral capacity emerges, and how it varies across rank measures.

Specifically, we sweep $\alpha \in \{0.5,1,1.5,2,3,5\}$ and measure \vspace{-1em}
\[
    R_\alpha^{\mathrm{pre}}, \qquad
    R_\alpha^{\mathrm{post}}, \qquad
    \rho_\alpha =
    \frac{R_\alpha^{\mathrm{post}}}{R_\alpha^{\mathrm{pre}}},
\] 
where $R_\alpha^{\mathrm{pre}}$ is computed before the FFN nonlinearity,
$R_\alpha^{\mathrm{post}}$ is computed after the nonlinearity, and
$\rho_\alpha$ measures nonlinear reinjection ratio. This separates
the optimizer-induced geometry entering the nonlinearity from the
representation geometry after activation. We use GPT-2 350M models since
it is the largest model used in this work, however, we observe the same
qualitative trends at 160M scale.


\begin{figure}[htbp]
\centering
\subfloat[Pre-activation spectrum \label{subfig:pre_act_alpha_rank}]{
    \includegraphics[width=.33\textwidth]{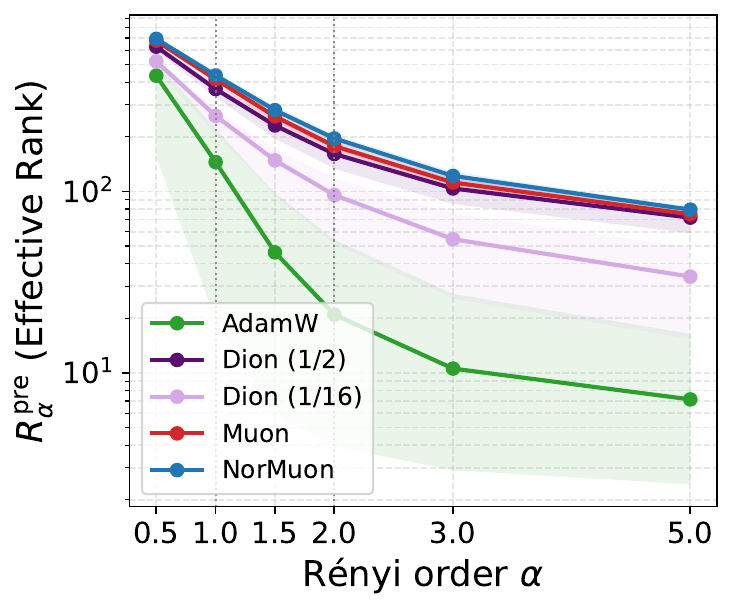}
}
\subfloat[Post-activation spectrum \label{subfig:post_act_alpha_rank}]{
    \includegraphics[width=.33\textwidth]{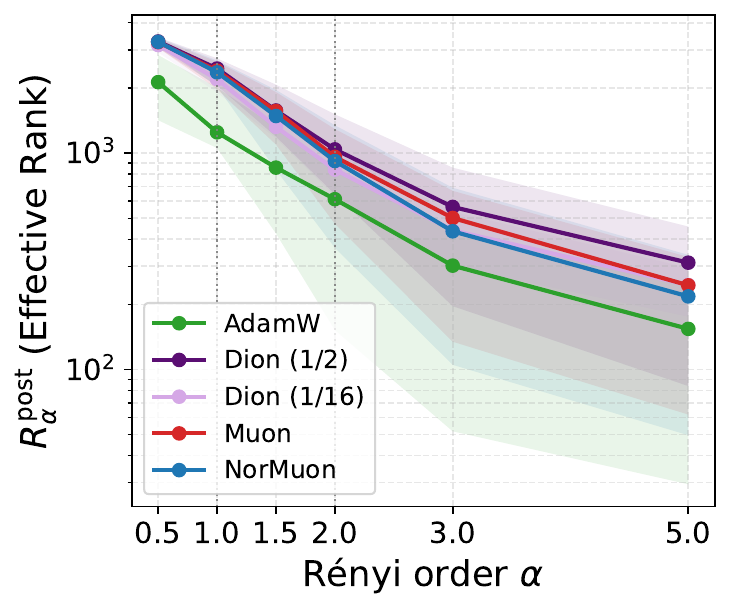}
}
\subfloat[Nonlinear reinjection \label{subfig:reinjection_alpha}]{
    \includegraphics[width=.33\textwidth]{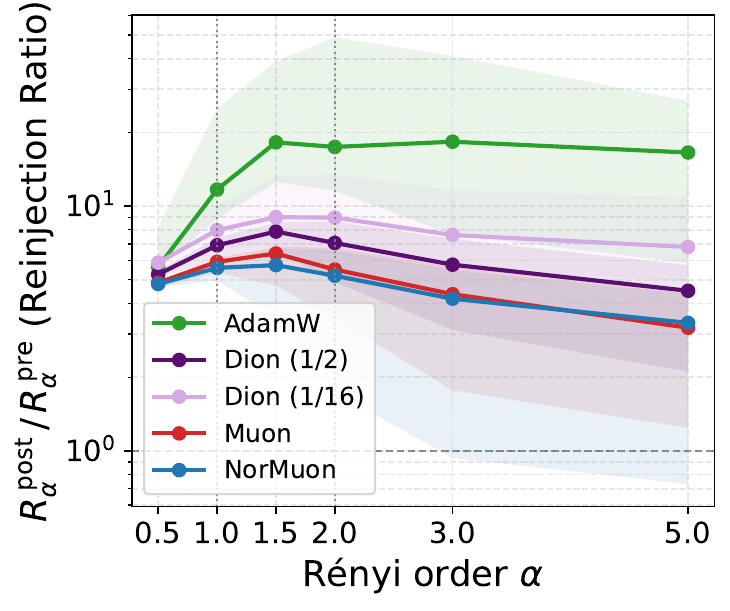}
} \vspace{-0.6em}
\caption{R\'enyi-family view of optimizer-shaped spectral capacity in GPT-2 350M,
with FFN width $4\times$. We report pre-activation rank
$R_\alpha^{\mathrm{pre}}$, post-activation rank
$R_\alpha^{\mathrm{post}}$, and nonlinear reinjection ratio
$\rho_\alpha = R_\alpha^{\mathrm{post}}/R_\alpha^{\mathrm{pre}}$.
Curves show the median across 24 layers, and shaded bands denote the interquartile range. Vertical guides mark
the soft-rank anchor $\alpha=1$ and the hard-rank anchor $\alpha=2$.}
\label{fig:renyi_family_350m}
\end{figure}

\paragraph{Pre-activation spectrum shows optimizer-induced geometry.}
Figure~\ref{fig:renyi_family_350m} shows the R\'enyi-family profiles for a
GPT-2 350M, and we observe that the strongest optimizer separation
appears before the FFN nonlinearity. Across all tested R\'enyi orders, the
pre-activation rank follows a stable hierarchy:
\[
\text{NorMuon} > \text{Muon} > \text{Dion}(r=1/2)
> \text{Dion}(r=1/16) > \text{AdamW}.
\]
This hierarchy aligns with validation perplexity at width $4\times$
(Table~\ref{tab:renyi_ppl_alignment}).

At the soft-rank anchor $\alpha=1$, pre-activation rank increases from
$126.0$ for AdamW to $411.3$ for NorMuon. At the hard-rank anchor $\alpha=2$,
the same monotonic structure holds:
\[
R_2^{\mathrm{pre}} =
177.4,\;164.8,\;148.0,\;84.5,\;28.0
\]
for NorMuon, Muon, Dion~$(r=1/2)$, Dion~$(r=1/16)$, and AdamW, respectively.
Thus, Muon and NorMuon enter the FFN nonlinearity with substantially richer
pre-activation spectra than AdamW, indicating that optimizer-induced spectral
geometry is already visible before activation.

\paragraph{The FFN nonlinearity re-organize the optimizer hierarchy.}
The post-activation spectra do not simply preserve the pre-activation ordering.
Although NorMuon has the largest pre-activation rank across the R\'enyi family,
Dion~$(r=1/2)$ achieves the largest post-activation rank for $\alpha \geq 1$.
At the hard-rank anchor $\alpha=2$, the post-activation ranks are
\[
R_2^{\mathrm{post}} =
989.1,\;864.4,\;811.3,\;798.5,\;546.7
\]
for Dion~$(r=1/2)$, Muon, NorMuon, Dion~$(r=1/16)$, and AdamW, respectively.
Thus, post-activation effective rank is not determined by pre-activation rank
alone. The FFN nonlinearity redistributes variance in an optimizer-dependent
way, causing rank-order changes between the pre- and post-activation spectra.
In other words, pre-activation ranks capture the
optimizer-induced precursor geometry, while post-activation ranks capture the
effective latent capacity available to subsequent layers in Transformer.  


\begin{wraptable}{r}{0.48\textwidth}
\vspace{-\intextsep}
\centering
\caption{Pre-activation rank and validation perplexity for GPT-2 350M at FFN
width $4\times$. Lower PPL is better, while higher pre-activation rank indicates
larger effective spectral capacity before the FFN nonlinearity.}
\label{tab:renyi_ppl_alignment}
\small
\setlength{\tabcolsep}{3.5pt}
\renewcommand{\arraystretch}{1.05}
\begin{tabular}{lccccc}
\toprule
Optimizer & PPL $\downarrow$
& $R_1^{\mathrm{pre}}$ & $R_2^{\mathrm{pre}}$
& $\rho_1$ & $\rho_2$ \\
\midrule
AdamW             & 28.35 & 126.0 & 28.0  & 22.6 & 51.9 \\
Dion~($r=1/16$)   & 24.92 & 238.0 & 84.5  & 8.9  & 12.4 \\
Dion~($r=1/2$)    & 23.27 & 344.3 & 148.0 & 6.5  & 6.4  \\
Muon              & 22.97 & 389.1 & 164.8 & 5.2  & 4.8  \\
NorMuon           & 22.80 & 411.3 & 177.4 & 5.2  & 4.3  \\
\bottomrule
\end{tabular}
\vspace{-1.0em}
\end{wraptable}
\paragraph{Large nonlinear reinjection is compensatory.}
The reinjection ratio $\rho_\alpha$ explains the post-activation reordering.
AdamW exhibits the largest reinjection across the R\'enyi family, peaking at
the hard-rank anchor with $\rho_2=51.9$. In contrast, Muon and NorMuon require
much smaller reinjection: $\rho_2=4.8$ for Muon and $\rho_2=4.3$ for NorMuon.
Thus, at $\alpha=2$, AdamW's reinjection is roughly $10.8\times$ larger than
Muon's and $12.1\times$ larger than NorMuon's. However, AdamW still has the
lowest post-activation $R_2$ and the worst validation perplexity. 
Thus, large reinjection does not implies 
better spectral capacity; rather, it indicates that the FFN nonlinearity is
compensating for a more collapsed optimizer-shaped precursor.

Dion further illustrates this effect as a controllable update-rank intervention.
At $\alpha=2$, decreasing the Dion rank fraction from $r=1/2$ to $r=1/16$
reduces $R_2^{\mathrm{pre}}$ from $148.0$ to $84.5$, while increasing the
reinjection ratio from $6.4$ to $12.4$, and their post-activation rank reduce from $989.1$ to $798.5$. 
Thus, aggressive low-rank updates reduce
the richness of the pre-activation geometry, but the FFN nonlinearity partially
recovers post-activation capacity.

This effect becomes especially visible in the high-$\alpha$ regime. At
$\alpha=5$, Dion~$(r=1/16)$ has much lower pre-activation rank than Muon and
NorMuon ($31.8$ versus $68.3$ and $71.4$), however its post-activation rank
slightly exceeds both ($221.7$ versus $217.3$ and $198.7$). Thus, {\em R\'enyi sweep
reveals  rank-order reversals that would be hidden by a single
effective-rank metric}.

Overall, the R\'enyi-family diagnostics separate three roles inside the FFN
block. Pre-activation spectra reveal the optimizer-shaped geometry entering the
nonlinearity; post-activation spectra show the representation passed to the FFN
output projection; and $\rho_\alpha$ measures how strongly the nonlinearity
reshapes the spectrum. In the 350M comparison, the best-performing optimizers
enter the nonlinearity with high pre-activation effective rank and require only
moderate reinjection. This links optimizer geometry to the spectral scaling laws
and suggests that R\'enyi-family profiles can serve
as useful diagnostics for optimizer--architecture interaction.


\section{Layer-Wise Robustness of Optimizer-Induced Spectral Scaling}
\label{app:layerwise_beta}

The main results in Section~\ref{sec:results:scaling} fit scaling exponents
from layer-aggregated spectral ranks. Here, we investigate whether those aggregate
trends reflect broad behavior across depth, or they are driven by a small number of
layers. For each layer $\ell$, optimizer, frequency regimes, and spectral rank metric, we
independently fit $R_\ell(D)\propto D^{\beta_\ell}$
across the same FFN widths used in the aggregate scaling-laws. This
produces one scaling exponent $\beta_\ell$ per layer. We summarize the
layer-wise distributions using the median, interquartile range (IQR), and the
fraction of layers with positive scaling exponent. A high positive-layer
fraction indicates that the aggregate scaling trend is broadly reflected across
depth.

\begin{figure}[t]
    \centering
    \includegraphics[width=0.98\textwidth]{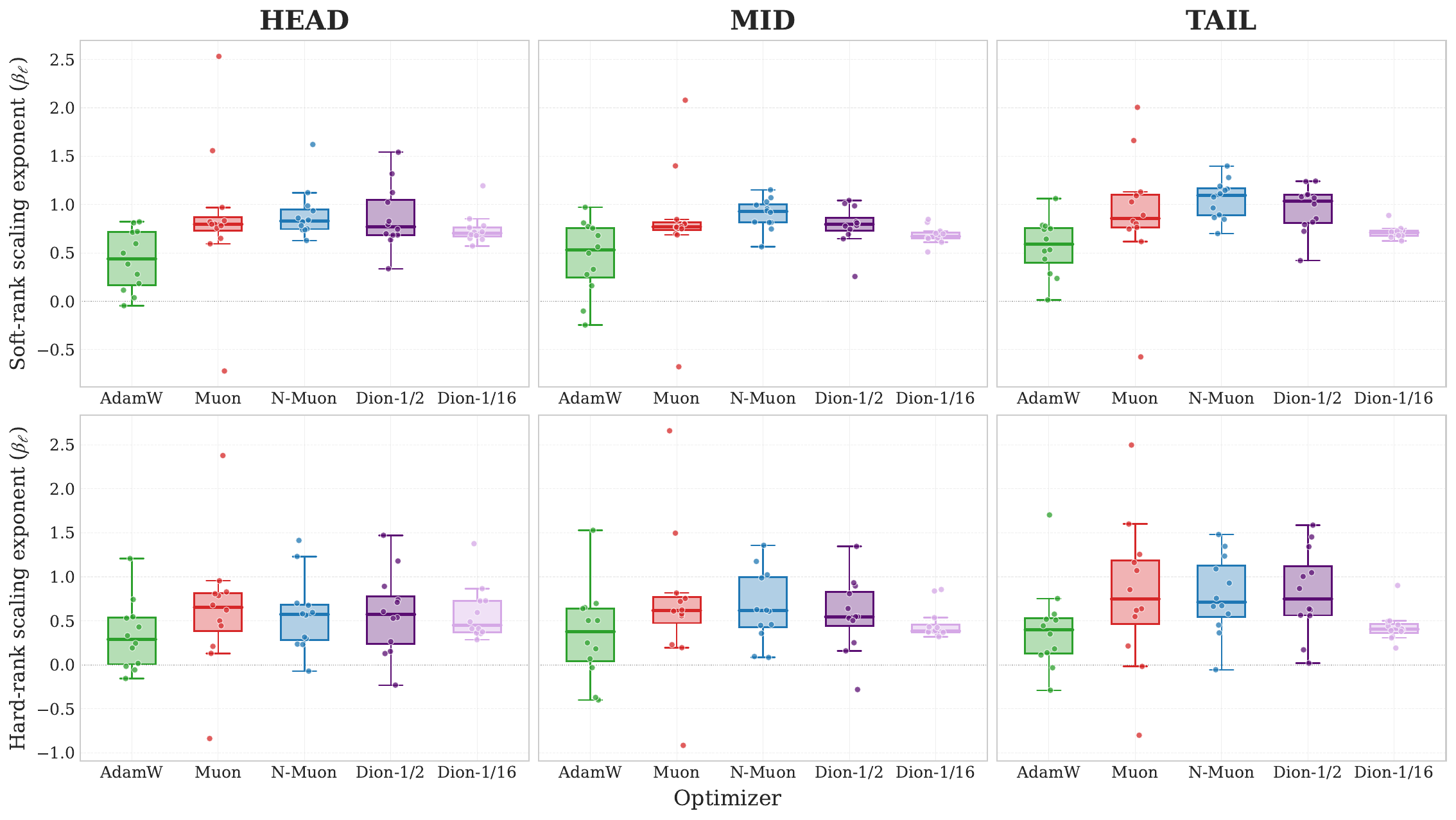} \vspace{-0.6em}
    \caption{Distribution of layer-wise scaling exponents for GPT-2 160M. For each
layer $\ell$, optimizer, token-frequency regimes, and spectral rank metric, we fit
$R_\ell(D)\propto D^{\beta_\ell}$ across FFN widths. Box plots summarize the
distribution of $\beta_\ell$ across layers, and the horizontal dotted line
marks $\beta_\ell=0$.}
    \label{fig:layerwise_beta_distribution_160m}
\end{figure}

\begin{figure}[t]
    \centering
    \includegraphics[width=0.98\textwidth]{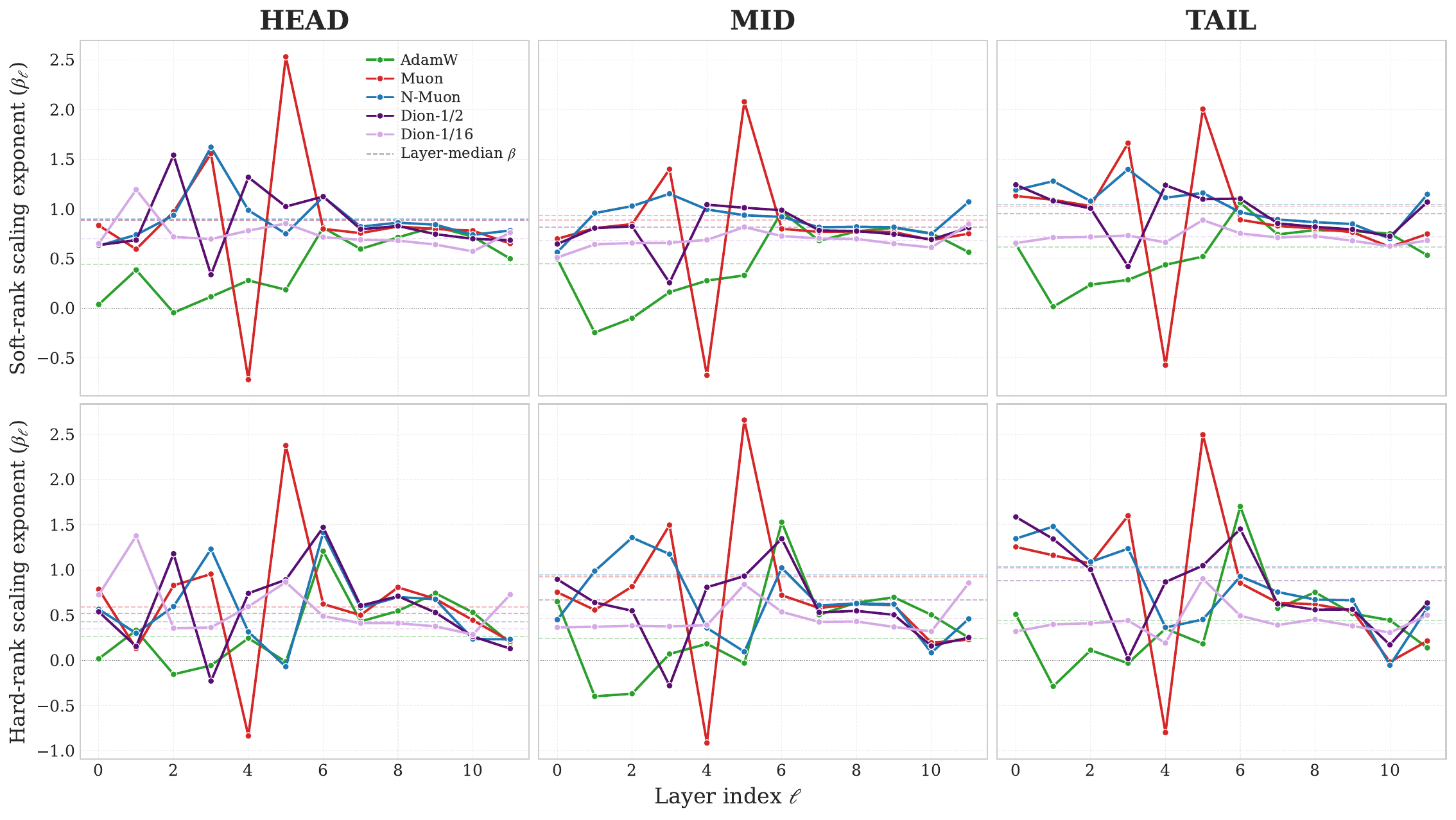} \vspace{-0.6em}
\caption{Depth profiles of layer-wise scaling exponents for GPT-2 160M. Each
curve shows $\beta_\ell$ as a function of layer index for a fixed optimizer,
frequency regimes, and rank metric. Dashed horizontal lines denote the
corresponding aggregate exponent from the main analysis.}
    \label{fig:layerwise_beta_depth_160m}
\end{figure}

Figure~\ref{fig:layerwise_beta_distribution_160m} shows the distribution of
layer-wise exponents, and Figure~\ref{fig:layerwise_beta_depth_160m} shows
their depth profiles. Tables~\ref{tab:layerwise_soft_beta}
and~\ref{tab:layerwise_hard_beta} report the corresponding robust summary
statistics. These diagnostics demonstrate the spread of layer-wise exponents and the
extent to which aggregate trends are reflected across depth.

\begin{table}[h]
\centering
\caption{Layer-wise soft-rank scaling exponents. For each
optimizer and token-frequency regimes, we report the median, interquartile range
(IQR), and fraction of layers with positive $\beta_\ell$.}
\label{tab:layerwise_soft_beta}
\resizebox{0.99\columnwidth}{!}{%
\begin{tabular}{lccc ccc ccc}
\toprule
& \multicolumn{3}{c}{HEAD}
& \multicolumn{3}{c}{MID}
& \multicolumn{3}{c}{TAIL} \\
\cmidrule(lr){2-4}\cmidrule(lr){5-7}\cmidrule(lr){8-10}
Optimizer
& Med. & IQR & Frac.$>0$
& Med. & IQR & Frac.$>0$
& Med. & IQR & Frac.$>0$ \\
\midrule
AdamW        & 0.441 & 0.547 & 0.917 & 0.530 & 0.511 & 0.833 & 0.588 & 0.358 & 1.000 \\
Muon         & 0.797 & 0.138 & 0.917 & 0.770 & 0.079 & 0.917 & 0.859 & 0.340 & 0.917 \\
NorMuon      & 0.831 & 0.201 & 1.000 & 0.927 & 0.188 & 1.000 & 1.095 & 0.281 & 1.000 \\
Dion (1/2)   & 0.770 & 0.363 & 1.000 & 0.794 & 0.133 & 1.000 & 1.036 & 0.286 & 1.000 \\
Dion (1/16)  & 0.705 & 0.093 & 1.000 & 0.673 & 0.061 & 1.000 & 0.711 & 0.053 & 1.000 \\
\bottomrule
\end{tabular}}
\end{table}

\paragraph{Added width broadly expands soft-rank capacity across depth.}
A positive layer-wise exponent $\beta_\ell>0$ implies  that the effective capacity of
layer $\ell$ increases with added FFN width. More importantly, it show
that the near-linear aggregate $\beta_{\mathrm{soft}}$ is not produced by a
small subset of layers. Soft-rank exponents are positive for nearly all layers,
and their medians remain large for Muon-style optimizers. NorMuon attains the
largest median soft-rank exponents across all regimes, reaching
$0.831$ in HEAD, $0.927$ in MID, and $1.095$ in TAIL. AdamW also shows positive
scaling in most layers, {\em but with weaker medians and wider layer-wise spread}.
Dion~$(1/16)$ is more constrained but highly stable, with small IQRs across
MID and TAIL regimes (Table \ref{tab:layerwise_soft_beta}). 
Thus, aggregate $\beta_{\mathrm{soft}}$ trends in the scaling laws
are not only directionally positive across depth, their magnitude and
optimizer ordering are also reflected in the layer-wise fits.

\paragraph{Hard-rank scaling is more heterogeneous but remains optimizer-dependent.}
Hard-rank exponents show stronger variation across depth than soft-rank
exponents, consistent with dominant-mode capacity being a more selective
spectral notion. The optimizer-dependent structure observed in the aggregate
fits remains visible at the layer level. For AdamW, $75\%$ of layers have
positive hard-rank scaling in HEAD and MID, and its median hard-rank exponents
remain below $0.40$ in all regimes. Muon, NorMuon, and Dion~$(1/2)$ produce
substantially larger median hard-rank exponents, although with greater
layer-wise spread than in the soft-rank case. The separation is clearest in
TAIL, where their median exponents are $0.746$, $0.714$, and $0.751$,
respectively (Table \ref{tab:layerwise_hard_beta}). 
This reinforce the aggregate scaling laws result that orthonormal-update optimizers
convert added FFN width into dominant-mode capacity more effectively than AdamW.

\begin{table}[h]
\centering
\caption{Layer-wise hard-rank scaling exponents. For each
optimizer and frequency regime, we report the median, interquartile range
(IQR), and fraction of layers with positive $\beta_\ell$.}
\label{tab:layerwise_hard_beta}
\resizebox{0.99\columnwidth}{!}{%
\begin{tabular}{lccc ccc ccc}
\toprule
& \multicolumn{3}{c}{HEAD}
& \multicolumn{3}{c}{MID}
& \multicolumn{3}{c}{TAIL} \\
\cmidrule(lr){2-4}\cmidrule(lr){5-7}\cmidrule(lr){8-10}
Optimizer
& Med. & IQR & Frac.$>0$
& Med. & IQR & Frac.$>0$
& Med. & IQR & Frac.$>0$ \\
\midrule
AdamW        & 0.288 & 0.525 & 0.750 & 0.377 & 0.597 & 0.750 & 0.397 & 0.400 & 0.833 \\
Muon         & 0.651 & 0.427 & 0.917 & 0.617 & 0.294 & 0.917 & 0.746 & 0.720 & 0.833 \\
NorMuon      & 0.573 & 0.398 & 0.917 & 0.615 & 0.569 & 1.000 & 0.714 & 0.579 & 0.917 \\
Dion (1/2)   & 0.571 & 0.543 & 0.917 & 0.548 & 0.390 & 0.917 & 0.751 & 0.559 & 1.000 \\
Dion (1/16)  & 0.450 & 0.353 & 1.000 & 0.386 & 0.086 & 1.000 & 0.404 & 0.098 & 1.000 \\
\bottomrule
\end{tabular}}
\end{table}

\paragraph{Low-rank Dion is stable but capacity-limited.}
The layer-wise analysis also clarifies the behavior of Dion~$(1/16)$. This
optimizer has positive hard-rank scaling in every layer across all 
regimes, with especially narrow IQRs in MID and TAIL (Table \ref{tab:layerwise_hard_beta}). 
However, its median
hard-rank exponents are much lower than those of Muon, NorMuon, and
Dion~$(1/2)$. For instance, in TAIL, Dion~$(1/16)$ has median
$\beta_\ell=0.404$, compared with $0.746$ for Muon, $0.714$ for NorMuon, and
$0.751$ for Dion~$(1/2)$. Thus, aggressive low-rank update structure yields
stable but constrained hard-rank scaling.

\paragraph{Depth profiles reveal structured heterogeneity.}
The depth profiles in Figure~\ref{fig:layerwise_beta_depth_160m} show that
layer-wise exponents vary across depth, especially for hard rank. This
variation helps explain why some aggregate hard-rank fits have lower $R^2$ in
the main table. Even with this heterogeneity, the optimizer ordering remains
visible across many layers: AdamW generally exhibits weaker hard-rank scaling,
whereas Muon, NorMuon, and Dion~$(1/2)$ frequently produce larger positive
exponents. These layer-wise fits support the aggregate scaling laws trends. 
Thus, optimizer-induced scaling differences are not artifacts of layer
aggregation; rather, {\em hard-rank capacity is more depth-sensitive and
optimizer-sensitive than soft-rank spread}.


\section{Hard-Rank Dynamics Under Extended AdamW Training}
\label{app:adamw_pr_dynamics}

Section~\ref{sec:results:matched_loss} shows that extended AdamW training
breaks the power-law relationship between FFN width and hard rank, {\em even when
validation perplexity continues to improve}. In this section, we investigate this through 
the post-activation hard-rank trajectories. We focus on
hard rank because the beta-dynamics analysis in Figure \ref{fig:beta_dynamics} shows 
the diminishing return in $\beta_{\mathrm{hard}}$(TAIL) for AdamW extended training. 
For a clean scaling law, larger FFN widths should preserve a systematic
width-capacity ordering. Extended AdamW training disrupts this ordering. For example, the $8d$ model
initially gains hard-rank capacity, but later drops below narrower models in
some frequency regimes. As a result, larger FFN width no longer corresponds to
larger hard-rank capacity, breaking the width--capacity relationship required
for reliable power-law fits.

\begin{figure}[h]
    \centering
    \includegraphics[width=\textwidth]{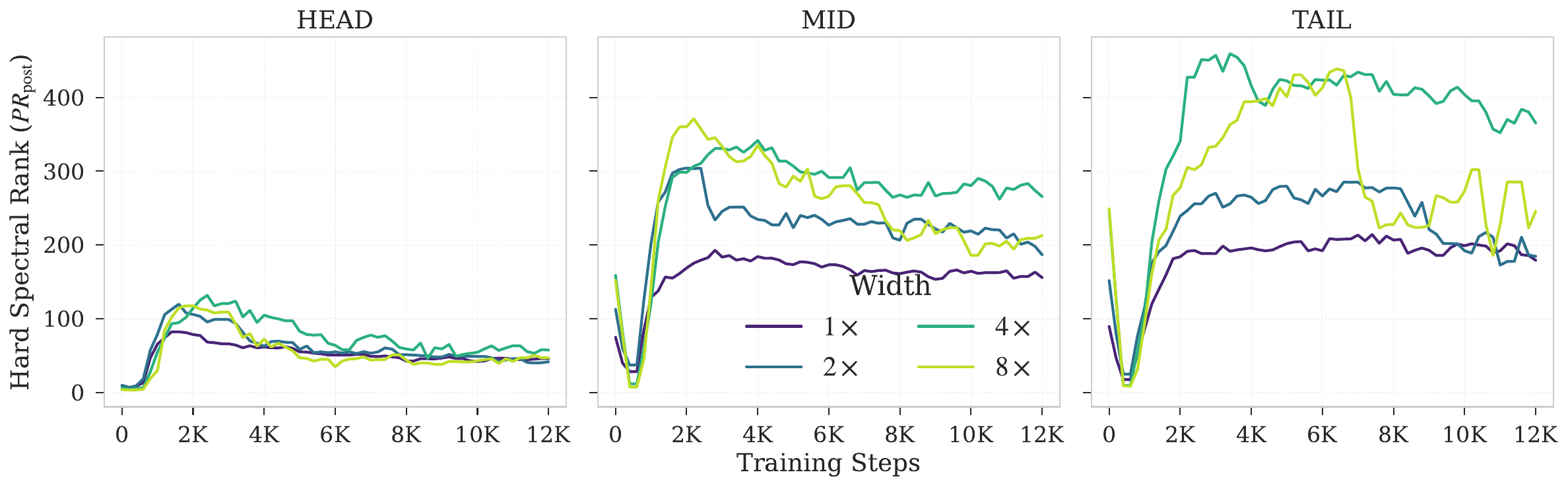}
    \caption{Hard-rank dynamics for AdamW extended training with GPT-2
    160M. Post-activation hard ranks are shown across 12K training steps for
    representative FFN widths ($1\times$, $2\times$, $4\times$, $8\times$). 
    In HEAD regime, all widths rise early and then
    decline toward similar values. In MID regimes, width ordering is partially violated
    at the end of training---$8\times$ trajectory falling below the
    $4\times$ trajectory. In TAIL regimes, the breakdown is {\bf strongest:} the $8\times$
    trajectory peaks early, drops sharply after roughly 7K steps, falls below
    the $2\times$ trajectory around 8K, and remains far below the $4\times$
    trajectory at 12K.}
    \label{fig:pr_dynamics_appendix}
\end{figure}

\paragraph{Differential hard-rank erosion explains the fit breakdown.}
Figure~\ref{fig:pr_dynamics_appendix} and Table~\ref{tab:pr_trajectories} 
 show that the hard-rank scaling failure, induced by a breakdown of width--capacity ordering. 
In the TAIL regimes, hard rank initially increases with width, consistent with a width-scaling relation.
During extended training, however, wider FFN configurations can lose
dominant-mode capacity more rapidly than narrower ones. The $8d$ model rises
from $278$ at 2K to $394$ at 4K, but then drops to $228$ by 8K, falling below
the $2d$ model at $277$. By 12K, the $1d$ and $2d$ models have nearly converged
($179$ vs. $185$), whereas the $4d$ model remains substantially higher
($365$). Thus, the hard-rank fit does not merely become noisy, the 
relationship between width and dominant-mode capacity is itself disrupted.

\begin{table}[h]
    \centering
    \caption{TAIL hard-rank dynamics under extended AdamW
    training for GPT-2 160M. Post-activation hard rank is reported at selected
    checkpoints for representative FFN widths. The $8d$ trajectory rises early
    but drops below the $2d$ model by 8K steps and remains far below the $4d$
    model at 12K, disrupting the width--capacity relationship needed for a
    reliable power-law fit.}
    \label{tab:pr_trajectories}
    \small
    \setlength{\tabcolsep}{3pt}
    \resizebox{\textwidth}{!}{%
    \begin{tabular}{l cccc cccc cccc cccc}
    \toprule
     & \multicolumn{4}{c}{$1d$}
     & \multicolumn{4}{c}{$2d$}
     & \multicolumn{4}{c}{$4d$}
     & \multicolumn{4}{c}{$8d$} \\
    \cmidrule(lr){2-5}
    \cmidrule(lr){6-9}
    \cmidrule(lr){10-13}
    \cmidrule(lr){14-17}
     & 2K & 4K & 8K & 12K
     & 2K & 4K & 8K & 12K
     & 2K & 4K & 8K & 12K
     & 2K & 4K & 8K & 12K \\
    \midrule
    Hard-rank
    & 184 & 196 & 207 & 179
    & 239 & 265 & 277 & 185
    & 341 & 416 & 404 & 365
    & 278 & 394 & 228 & 246 \\
    \bottomrule
    \end{tabular}%
    }
\end{table}

The same pattern appears, more mildly, outside the TAIL regime. In HEAD regime, the
$8\times$ trajectory declines from roughly $117$ near its early peak to about
$47$ by 12K, while the smaller-width trajectories converge toward similar
values. In MID regime, the $8\times$ trajectory falls below the $4\times$ trajectory
by the end of training. These dynamics support the interpretation in
Section~\ref{sec:results:matched_loss}, extended AdamW training does not shift
all widths uniformly, but preferentially erodes hard-rank capacity in wider FFN
configurations. This breaks the width--capacity ordering needed for hard-rank scaling, 
{\bf despite the continuous improvement in validation perplexity.}


\section{The AdamW--Muon Spectral-Scaling Gap Persists Across Learning Rates}
\label{app:lr_ablation}

A natural concern is that the spectral-scaling gap between AdamW and Muon could
be the artifact of  learning-rate tuning rather than optimizer geometry. We investigate this by
sweeping learning rates for both optimizers and analyzing  their spectral scaling behavior. 
AdamW is evaluated at $\{10^{-3},3\times10^{-3},6\times10^{-3}\}$, with
$3\times10^{-3}$ used as the default in the main experiments. Muon is evaluated
at $\{10^{-2},2\times10^{-2},4\times10^{-2}\}$, with
$2\times10^{-2}$ used as the default. This sweep is intended as a scalar
learning-rate control, rather than an exhaustive hyperparameter search.

\paragraph{Learning-rate tuning does not recover Muon-like hard-rank scaling.}
Table~\ref{tab:lr_ablation} shows that the AdamW--Muon gap is not closed by the
tested learning-rate sweep. The clearest comparison is TAIL hard rank, AdamW's
largest reliable exponent is $\beta=0.44$ at $3\times10^{-3}$, whereas Muon's
lowest exponent across the sweep is $\beta=0.80$ at $10^{-2}$. Thus, even
Muon's weakest learning-rate setting exceeds AdamW's strongest reliable setting
by $0.36$. Moreover, this gap is larger than the within-AdamW reliable
range ($0.44-0.32=0.12$) and the within-Muon range ($1.02-0.80=0.22$). That is, the
TAIL hard-rank exponent ranges do not overlap across the tested learning
rates. AdamW learning-rate sweep redistribute weak scaling across
frequency regimes, at $10^{-3}$, $\beta_{\mathrm{hard}}$(HEAD) improves to $0.43$ but
MID collapses to $\beta$=0.00. At $3\times10^{-3}$, $\beta_{\mathrm{hard}}$(TAIL) improves to $0.44$
while HEAD and MID remain weak, and  at $6\times10^{-3}$, the TAIL hard-rank fit
becomes unreliable ($R^2=0.04$).

\begin{table}[h]
    \centering
    \caption{Spectral scaling exponents across learning rates for GPT-2 160M.
    Hard-rank and soft-rank scaling exponents $\beta$ are reported across
    token-frequency regimes, with $R^2$ in parentheses.}
    \label{tab:lr_ablation}
    \resizebox{\columnwidth}{!}{%
    \begin{tabular}{l c ccc ccc}
    \toprule
    & & \multicolumn{3}{c}{Hard Rank} & \multicolumn{3}{c}{Soft Rank} \\
    \cmidrule(lr){3-5} \cmidrule(lr){6-8}
    Optimizer & LR & HEAD & MID & TAIL & HEAD & MID & TAIL \\
    \midrule
    AdamW & $10^{-3}$
    & 0.43 \scriptsize{(0.57)}
    & 0.00 \scriptsize{(0.00)}
    & 0.32 \scriptsize{(0.64)}
    & 0.63 \scriptsize{(0.89)}
    & 0.52 \scriptsize{(0.76)}
    & 0.70 \scriptsize{(0.99)} \\
    AdamW & $3\times10^{-3}$
    & 0.26 \scriptsize{(0.59)}
    & 0.24 \scriptsize{(0.36)}
    & 0.44 \scriptsize{(0.66)}
    & 0.44 \scriptsize{(0.82)}
    & 0.45 \scriptsize{(0.82)}
    & 0.62 \scriptsize{(0.97)} \\
    AdamW & $6\times10^{-3}$
    & 0.25 \scriptsize{(0.19)}
    & 0.28 \scriptsize{(0.29)}
    & 0.12 \scriptsize{(0.04)}
    & 0.43 \scriptsize{(0.47)}
    & 0.36 \scriptsize{(0.43)}
    & 0.34 \scriptsize{(0.33)} \\
    \midrule
    Muon & $10^{-2}$
    & 0.75 \scriptsize{(0.90)}
    & 0.77 \scriptsize{(0.87)}
    & 0.80 \scriptsize{(0.97)}
    & 0.99 \scriptsize{(0.99)}
    & 0.85 \scriptsize{(1.00)}
    & 0.91 \scriptsize{(0.99)} \\
    Muon & $2\times10^{-2}$
    & 0.59 \scriptsize{(0.54)}
    & 0.93 \scriptsize{(0.82)}
    & 1.02 \scriptsize{(0.81)}
    & 0.88 \scriptsize{(0.90)}
    & 0.89 \scriptsize{(0.96)}
    & 1.03 \scriptsize{(0.94)} \\
    Muon & $4\times10^{-2}$
    & 0.69 \scriptsize{(0.43)}
    & 0.71 \scriptsize{(0.82)}
    & 0.96 \scriptsize{(0.93)}
    & 0.82 \scriptsize{(0.94)}
    & 0.83 \scriptsize{(0.97)}
    & 0.97 \scriptsize{(0.99)} \\
    \bottomrule
    \end{tabular}%
    }
\end{table}

\paragraph{Soft-rank scaling shows the same ordering with a smaller gap.}
The same qualitative separation appears for $\beta_{\mathrm{soft}}$, although
the gap is smaller than $\beta_{\mathrm{hard}}$. AdamW's largest TAIL soft-rank exponent
is $\beta=0.70$ at $10^{-3}$, whereas Muon's lowest TAIL soft-rank exponent is
$\beta=0.91$ at $10^{-2}$. The ordering remains unchanged across the tested
learning rates. Soft-rank fits are also more stable for Muon, with TAIL
$R^2\geq0.94$ across the sweep.


Thus, the above learning-rate sweep ablations rule out a simple scalar
learning-rate explanation for the AdamW--Muon gap. Combined with the
extended-training control in Appendix~\ref{app:adamw_pr_dynamics}, this shows
that AdamW's weaker hard-rank scaling is not explained by insufficient training
time or learning-rate mistuning. This supports the interpretation that optimizer
geometry changes how added FFN width is converted into usable capacity.


\section{Scaling Exponents for Dion Rank Sweep} 
\label{app:dion_rank_sweep_betas}

Section~\ref{subsec:dion_rank_sweep} shows that reducing Dion's update rank
lowers hard-capacity scaling, especially for TAIL tokens. Table~\ref{tab:dion_rank_sweep_betas}
reports the full fitted exponents underlying that comparison. The clearest
trend appears in MID and TAIL, where decreasing the rank fraction from
$r=1/2$ to $r=1/16$ reduces hard-rank scaling more sharply than soft-rank
scaling. For instance, $\beta_{\mathrm{hard}}$(TAIL) drops from $0.88$ to
$0.40$, while $\beta_{\mathrm{soft}}$ decreases gradually from $0.95$ to
$0.72$. HEAD trends are less monotonic and have larger uncertainty, so the
rank-bottleneck interpretation is most reliable in the MID and TAIL token regimes.

\begin{table}[h]
\centering
\setlength{\tabcolsep}{6pt}
\caption{Scaling exponents across Dion rank fractions for GPT-2 160M.
For each optimizer, token-frequency regimes, and rank metric, we report the
fitted exponent $\beta$ with fit uncertainty and $R^2$ in parentheses.
Decreasing Dion's rank fraction primarily reduces hard-rank scaling, especially
in MID and TAIL, while soft-rank scaling degrades more gradually.}
\label{tab:dion_rank_sweep_betas}
\resizebox{\textwidth}{!}{%
\begin{tabular}{lcccccc}
\toprule
 & \multicolumn{2}{c}{HEAD} & \multicolumn{2}{c}{MID} & \multicolumn{2}{c}{TAIL} \\
\cmidrule(lr){2-3} \cmidrule(lr){4-5} \cmidrule(lr){6-7}
Optimizers & Hard Rank & Soft Rank & Hard Rank & Soft Rank & Hard Rank & Soft Rank \\
\midrule
AdamW
 & \makecell{$0.26 \pm 0.22$ \\ ($R^2 = 0.59$)}
 & \makecell{$0.44 \pm 0.21$ \\ ($R^2 = 0.82$)}
 & \makecell{$0.24 \pm 0.32$ \\ ($R^2 = 0.36$)}
 & \makecell{$0.45 \pm 0.21$ \\ ($R^2 = 0.82$)}
 & \makecell{$0.44 \pm 0.31$ \\ ($R^2 = 0.66$)}
 & \makecell{$0.62 \pm 0.10$ \\ ($R^2 = 0.97$)} \\
Dion (1/2)
 & \makecell{$0.52 \pm 0.60$ \\ ($R^2 = 0.43$)}
 & \makecell{$0.89 \pm 0.30$ \\ ($R^2 = 0.90$)}
 & \makecell{$0.67 \pm 0.10$ \\ ($R^2 = 0.98$)}
 & \makecell{$0.82 \pm 0.06$ \\ ($R^2 = 0.99$)}
 & \makecell{$0.88 \pm 0.13$ \\ ($R^2 = 0.98$)}
 & \makecell{$0.95 \pm 0.07$ \\ ($R^2 = 1.00$)} \\
Dion (1/4)
 & \makecell{$0.72 \pm 0.34$ \\ ($R^2 = 0.82$)}
 & \makecell{$0.90 \pm 0.16$ \\ ($R^2 = 0.97$)}
 & \makecell{$0.58 \pm 0.06$ \\ ($R^2 = 0.99$)}
 & \makecell{$0.77 \pm 0.03$ \\ ($R^2 = 1.00$)}
 & \makecell{$0.67 \pm 0.06$ \\ ($R^2 = 0.99$)}
 & \makecell{$0.86 \pm 0.03$ \\ ($R^2 = 1.00$)} \\
Dion (1/8)
 & \makecell{$0.56 \pm 0.27$ \\ ($R^2 = 0.81$)}
 & \makecell{$0.75 \pm 0.16$ \\ ($R^2 = 0.96$)}
 & \makecell{$0.53 \pm 0.08$ \\ ($R^2 = 0.98$)}
 & \makecell{$0.74 \pm 0.03$ \\ ($R^2 = 1.00$)}
 & \makecell{$0.48 \pm 0.04$ \\ ($R^2 = 1.00$)}
 & \makecell{$0.78 \pm 0.02$ \\ ($R^2 = 1.00$)} \\
Dion (1/16)
 & \makecell{$0.35 \pm 0.20$ \\ ($R^2 = 0.75$)}
 & \makecell{$0.70 \pm 0.09$ \\ ($R^2 = 0.98$)}
 & \makecell{$0.46 \pm 0.07$ \\ ($R^2 = 0.98$)}
 & \makecell{$0.68 \pm 0.05$ \\ ($R^2 = 1.00$)}
 & \makecell{$0.40 \pm 0.04$ \\ ($R^2 = 0.99$)}
 & \makecell{$0.72 \pm 0.03$ \\ ($R^2 = 1.00$)} \\
\bottomrule
\end{tabular}}
\end{table}


\section{Attention-Rank Effects and Optimizer--Architecture Interactions}
\label{app:optimizer_architecture_effects}

Section~\ref{sec:results:optimizer_vs_architecture} compares optimizer-induced
spectral-scaling gains with the effect of architectural intervention, attention-rank, reducing the number of attention
heads from $12$ to $6$ at fixed parameter count. Here, we report the exact effect sizes, 
and interaction diagnostics for both, the optimizer-induced and architecture -induced spectral scaling shift. In particular, the
goal is to separate three quantities: (i) the optimizer-induced gain over AdamW, (ii) the
absolute attention-rank architectural shift, and (iii) the change in optimizer gain
induced by the architectural intervention. 

Table~\ref{tab:optimizer_vs_arch_effect} reports these quantities. 
The optimizer-induced gain exceeds the absolute attention-rank shift in $28$ of
$30$ regime--optimizer comparisons (Figure \ref{fig:optimizer_vs_architecture}). 
The only exceptions occur for HEAD hard-rank scaling under AdamW and Muon, where increasing per-head attention
rank produces a larger shift than the best optimizer-induced gain over AdamW.
Thus, across nearly all
regimes, changing optimizer produces a larger spectral-scaling shift than this
controlled attention-rank intervention.


\begin{table}[h]
\centering
\setlength{\tabcolsep}{4pt}
\caption{Optimizer-induced gains over AdamW and absolute attention-rank
architectural shifts for GPT-2 160M. $\Delta\beta_{\mathrm{opt}}^{\star}$
measures the best optimizer-induced gain over AdamW under the 12-head
architecture. Columns to the right report
$A_{\mathrm{arch}}(o)=|\beta_{o,6h}-\beta_{o,12h}|$ for each optimizer.}
\label{tab:optimizer_vs_arch_effect}
\begin{tabular}{lccccccc}
\toprule
& & & \multicolumn{5}{c}{$A_{\mathrm{arch}}(o)$} \\
\cmidrule(lr){4-8}
Regime & Best opt. & $\Delta\beta_{\mathrm{opt}}^{\star}$
& AdamW & Muon & NorMuon & Dion (1/2) & Dion (1/16) \\
\midrule
$\beta_{\mathrm{hard}}(\mathrm{HEAD})$ & Muon    & 0.330 & 0.345 & 0.651 & 0.060 & 0.130 & 0.039 \\
$\beta_{\mathrm{soft}}(\mathrm{HEAD})$ & NorMuon & 0.457 & 0.250 & 0.419 & 0.133 & 0.206 & 0.048 \\
$\beta_{\mathrm{hard}}(\mathrm{MID})$  & NorMuon & 0.703 & 0.429 & 0.294 & 0.131 & 0.086 & 0.093 \\
$\beta_{\mathrm{soft}}(\mathrm{MID})$  & NorMuon & 0.481 & 0.147 & 0.230 & 0.104 & 0.015 & 0.011 \\
$\beta_{\mathrm{hard}}(\mathrm{TAIL})$ & NorMuon & 0.600 & 0.313 & 0.367 & 0.023 & 0.014 & 0.011 \\
$\beta_{\mathrm{soft}}(\mathrm{TAIL})$ & NorMuon & 0.424 & 0.051 & 0.198 & 0.019 & 0.034 & 0.002 \\
\bottomrule
\end{tabular}
\end{table}

The architectural intervention also changes the optimizer gap itself. We
quantify this through
\begin{equation}
    I_{\star}^{(b,m)}
    =
    \Delta\beta_{\mathrm{opt}}^{\star,(b,m)}(6h)
    -
    \Delta\beta_{\mathrm{opt}}^{\star,(b,m)}(12h).
\end{equation}
Positive values indicate that reducing the number of heads increases the best
achievable optimizer-induced gain over AdamW. Table~\ref{tab:best_optimizer_interaction}
shows that $I_\star$ is positive in five of six regimes, with the largest
increase in HEAD hard-rank scaling ($+0.306$). The only negative interaction
occurs in MID hard rank, where the best gain decreases by $0.153$ but remains
large. Thus, the architectural intervention changes not only the scaling
exponents themselves, but also the size of the optimizer gap.

\begin{table}[h]
\centering
\setlength{\tabcolsep}{5.5pt}
\caption{Interaction of attention-rank architecture with the AdamW-to-best
optimizer gain for GPT-2 160M. We report the best optimizer-induced gain over
AdamW under the 12-head and 6-head architectures, together with
$I_{\star}=\Delta\beta_{\mathrm{opt}}^{\star}(6h)
-\Delta\beta_{\mathrm{opt}}^{\star}(12h)$.}
\label{tab:best_optimizer_interaction}
\begin{tabular}{lccc}
\toprule
Regime
& $\Delta\beta_{\mathrm{opt}}^{\star}(12h)$
& $\Delta\beta_{\mathrm{opt}}^{\star}(6h)$
& Interaction $I_{\star}$ \\
\midrule
$\beta_{\mathrm{hard}}(\mathrm{HEAD})$ & 0.330 & 0.636 & +0.306 \\
$\beta_{\mathrm{soft}}(\mathrm{HEAD})$ & 0.457 & 0.612 & +0.155 \\
$\beta_{\mathrm{hard}}(\mathrm{MID})$  & 0.703 & 0.550 & -0.153 \\
$\beta_{\mathrm{soft}}(\mathrm{MID})$  & 0.481 & 0.519 & +0.038 \\
$\beta_{\mathrm{hard}}(\mathrm{TAIL})$ & 0.600 & 0.637 & +0.037 \\
$\beta_{\mathrm{soft}}(\mathrm{TAIL})$ & 0.424 & 0.555 & +0.131 \\
\bottomrule
\end{tabular}
\end{table}

Table~\ref{tab:head_count_full_betas} reports the fitted exponents used to
compute both the signed architectural effects and the optimizer-induced gains.
These values show a {\bf stronger form of optimizer--architecture interaction}, the
attention-rank intervention changes which optimizer is best matched
to the architecture. Under the 12-head baseline, NorMuon attains the largest
scaling exponent in five of six regimes, with Muon leading only in HEAD hard
rank. Under the 6-head architecture, Muon attains the largest exponent in all
six regimes. Thus, reducing the number of heads does not act as a uniform
offset applied to all optimizers. It changes the optimizer ordering
itself.

\begin{table}[h]
\centering
\setlength{\tabcolsep}{4pt}
\caption{Fitted scaling exponents under 12-head and 6-head architectures for
GPT-2 160M. Each entry reports $(\beta_{12h},\beta_{6h})$ for the corresponding
optimizer, frequency regime, and rank metric.}
\label{tab:head_count_full_betas}
\begin{tabular}{lccccc}
\toprule
Regime & AdamW & Muon & NorMuon & Dion (1/2) & Dion (1/16) \\
\midrule
$\beta_{\mathrm{hard}}(\mathrm{HEAD})$ & (0.263, 0.608) & (0.593, 1.244) & (0.429, 0.489) & (0.520, 0.390) & (0.349, 0.388) \\
$\beta_{\mathrm{soft}}(\mathrm{HEAD})$ & (0.441, 0.691) & (0.884, 1.303) & (0.898, 0.765) & (0.886, 0.680) & (0.699, 0.651) \\
$\beta_{\mathrm{hard}}(\mathrm{MID})$  & (0.242, 0.671) & (0.927, 1.221) & (0.945, 1.076) & (0.666, 0.580) & (0.460, 0.367) \\
$\beta_{\mathrm{soft}}(\mathrm{MID})$  & (0.449, 0.596) & (0.885, 1.115) & (0.930, 1.034) & (0.815, 0.830) & (0.679, 0.668) \\
$\beta_{\mathrm{hard}}(\mathrm{TAIL})$ & (0.438, 0.751) & (1.021, 1.388) & (1.038, 1.015) & (0.879, 0.865) & (0.404, 0.393) \\
$\beta_{\mathrm{soft}}(\mathrm{TAIL})$ & (0.618, 0.669) & (1.026, 1.224) & (1.042, 1.061) & (0.952, 0.918) & (0.718, 0.716) \\
\bottomrule
\end{tabular}
\end{table}

{\bf These diagnostics support the arguments for architecture-optimizer co-design.}
Optimizer-induced gains are usually larger than the direct attention-rank-induced
shift, and the same architectural intervention has optimizer-dependent effects, 
increasing per-head attention rank changes which optimizer achieves the
largest spectral-scaling exponent. In other words, attention-rank changes and optimizer
geometry act as coupled, rather than separable, axes of
representation scaling.


\section{Optimizer Geometry Expands the Trainable Normalization Space}
\label{sec:results:mixln}

Sections~\ref{sec:results:optimizer_vs_architecture} and
Appendix~\ref{app:optimizer_architecture_effects} show that optimizer geometry
changes the  spectral scaling behavior of  trainable architectures. Here, we
study a complementary form of optimizer--architecture coupling: whether
optimizer choice changes which normalization-placement architectures can be
trained at useful perplexity in the first place. This is not a spectral-scaling
experiment as all models use the same $4\times$ FFN width. Instead, it tests
whether optimizer geometry changes the feasible region of the architecture
search space before spectral scaling is measured.

We use normalization placement as a trainability stress test. In a partial
PostLN configuration, denoted PostLN-$k$, the first $k\%$ of layers use PostLN,
while the remaining layers use PreLN. This creates a controlled interpolation
between the more trainable PreLN regime and the more difficult PostLN regime.
Since PostLN is known to be difficult to train at scale due to gradient
amplification across layers~\citep{xiong2020layer}, varying the PostLN fraction
lets us test whether optimizers differ in the normalization configurations they
can train to useful perplexity.

\begin{table}[t]
    \centering
    \caption{Validation perplexity for partial PostLN configurations in GPT-2
    160M at $4\times$ FFN width. In PostLN-$k$, the first $k\%$ of layers use
    PostLN and the remaining layers use PreLN. Diverged runs (PPL $>1000$) are
    marked with \xmark. Lower is better.}
    \label{tab:mixln}
    \begin{tabular}{l cccc}
    \toprule
    Optimizer & PostLN-25 & PostLN-50 & PostLN-75 & Full PostLN \\
    \midrule
    AdamW (lr=$10^{-4}$) & 64.6 & 65.6 & 106.7 & \xmark \\
    AdamW (lr=$3\times10^{-4}$) & 41.9 & \xmark & \xmark & \xmark \\
    \midrule
    Muon & 28.7 & 30.1 & 40.9 & \xmark \\
    NorMuon & 28.7 & 29.9 & \textbf{32.8} & \xmark \\
    Dion ($r=1/2$) & 29.1 & 31.7 & \xmark & \xmark \\
    Dion ($r=1/16$) & 30.7 & 34.7 & \xmark & \xmark \\
    \bottomrule
    \end{tabular}
\end{table}

\paragraph{Muon-family optimizers remain trainable in higher-PostLN regimes.}
Table~\ref{tab:mixln} shows a clear trainability gap. At
lr=$3\times10^{-4}$, AdamW trains PostLN-25 but diverges for PostLN-50 and
PostLN-75. Lowering the AdamW learning rate to $10^{-4}$ avoids divergence up
to PostLN-75, but it reaches to PPL $=106.7$,
compared with PPL $=40.9$ for Muon and PPL $=32.8$ for NorMuon. Thus, AdamW can
be made stable only by moving to a substantially worse optimization regime,
whereas Muon-family optimizers train these partial PostLN configurations at
useful perplexity.

\paragraph{Neuron-wise normalization helps in the most aggressive partial-PostLN regime.}
Muon and NorMuon perform similarly at lower PostLN fractions, both achieve
PPL $=28.7$ at PostLN-25 and nearly identical perplexity at PostLN-50. The
difference appears at PostLN-75, where NorMuon improves over Muon from
PPL $=40.9$ to PPL $=32.8$. This suggests that NorMuon's neuron-wise
normalization of orthogonalized updates provides additional stability when the
PostLN fraction is large and gradient amplification is more severe. However, this
stabilization is not unlimited, full PostLN remains unstable for all optimizers.

\paragraph{Trainability and spectral-capacity scaling are related but distinct.}
The Dion variants clarify that the trainability frontier is not determined only
by update rank. Both Dion(1/2) and Dion(1/16) train PostLN-50, but
both fail at PostLN-75. This contrasts with the spectral-scaling results in
Section~\ref{subsec:dion_rank_sweep}, where Dion rank strongly modulates
capacity exponents. This distinction suggests that orthonormal update structure
is important for stabilizing difficult normalization configurations, while
update rank controls how much of the available capacity is converted into
effective dominant-mode capacity.

These results broaden the co-design interpretation. Optimizer
choice changes how trainable architectures convert added FFN width into
useful spectral capacity. Here, optimizer choice changes which normalization-placement
architectures are trainable at useful perplexity in the first place. Thus,
optimizer geometry affects both {\em capacity utilization} within a fixed
architecture and the {\em feasible architecture space} itself.


\section{Position-Dependent FFN Spectral Transformation Under RoPE and NoPE}
\label{app:positional}

Section~\ref{sec:results:nope} shows that removing
RoPE induces optimizer-dependent spectral redistribution. In this section, we 
provide a position-dependent analysis, and investigate whether
the redistribution is accompanied by changes in how the FFN nonlinearity
transform position-dependent information. 

\paragraph{Symmetry-ratio metric.}
Following~\citep{lavie2024towards}, we quantify how much sequence position
explains FFN activation variance. For activations within a given layer and
token-frequency regime, we compute an ANOVA-style position-dependence score \vspace{-0.6em}
\begin{equation} 
    \eta^2
    =
    \frac{\mathrm{Var}_{\mathrm{between\ position}}}
         {\mathrm{Var}_{\mathrm{total}}}.
\end{equation}
We report its complement, the symmetry ratio \vspace{-0.6em}
\begin{equation}
    \mathrm{SR}=1-\eta^2.
\end{equation}
Higher $\mathrm{SR}$ means the representation is more position-independent,
lower $\mathrm{SR}$ means sequence position explains more activation variance.
We compute this statistic before and after the FFN nonlinearity:
$\mathrm{SR}_{\mathrm{pre}}$ and and $\mathrm{SR}_{\mathrm{post}}$. We define \vspace{-0.6em}
\begin{equation}
    \Delta\mathrm{SR}
    =
    \mathrm{SR}_{\mathrm{post}}-\mathrm{SR}_{\mathrm{pre}}.
\end{equation}
Thus, $\Delta\mathrm{SR}<0$ indicates that the FFN nonlinearity makes the
representation more position-dependent, while $\Delta\mathrm{SR}>0$ indicates
suppression of position-dependence. In the tables below,
$\overline{|\Delta\mathrm{SR}|}$ measures the layer-averaged magnitude of
FFN-induced positional processing, while the sign of $\Delta\mathrm{SR}$
determines whether this processing amplifies or suppresses position-dependence.
{\bf This analysis separates three effects:} whether NoPE changes the amount of
FFN-induced positional processing, where in depth this processing is localized,
and which token-frequency regime receives stronger amplification.


\subsection{Positional Amplification Tracks TAIL Hard-Rank Scaling}

Before analyzing RoPE/NoPE, we first check whether the symmetry-ratio
diagnostic tracks the TAIL hard-rank scaling behavior observed in the 160M
baseline. Table~\ref{tab:positional_160m} shows that {\em optimizers with stronger
TAIL hard-rank scaling also exhibit stronger TAIL positional amplification.}
NorMuon, Muon, and Dion~$(1/2)$ have the largest
$\overline{|\Delta\mathrm{SR}|}$ values and the largest TAIL hard-rank
exponents, whereas AdamW and Dion~$(1/16)$ show near-zero amplification and
weaker hard-rank scaling. This does not establish causality, but it suggests
that improved TAIL dominant-mode scaling is accompanied by stronger
FFN-mediated position-dependent processing.

\begin{table}[h]
    \centering
    \setlength{\tabcolsep}{5pt}
    \caption{TAIL positional amplification and hard-rank scaling at 160M. We
    report layer-averaged $\overline{|\Delta\mathrm{SR}|}$ for TAIL tokens, the
    most negative layer-wise $\Delta\mathrm{SR}$ value, its peak layer, and the
    corresponding TAIL hard-rank scaling exponent. More negative peak values
    indicate stronger FFN-induced amplification of position-dependence.}
    \label{tab:positional_160m}
    \begin{tabular}{l cccc}
    \toprule
    Optimizer
    & $\overline{|\Delta\mathrm{SR}|}$
    & Peak $\Delta\mathrm{SR}$
    & Peak Layer
    & $\beta_{\mathrm{hard}}$ \\
    \midrule
    NorMuon     & 0.087 & $-$0.711 & 5  & 1.04 \\
    Muon        & 0.078 & $-$0.529 & 4  & 1.02 \\
    Dion (1/2)  & 0.075 & $-$0.441 & 3  & 0.88 \\
    Dion (1/16) & 0.005 & $-$0.041 & 6  & 0.40 \\
    AdamW       & 0.003 & $-$0.008 & 10 & 0.44 \\
    \bottomrule
    \end{tabular}
\end{table}

\subsection{NoPE Changes the Amount of TAIL Positional Processing}

We next compare RoPE and NoPE at 350M. Table~\ref{tab:rope_nope_tail_abs}
summarizes the layer-averaged magnitude of TAIL positional processing at the
smallest and largest FFN widths in the sweep. Removing RoPE substantially
increases TAIL positional amplification for Muon and NorMuon. At $1d$, Muon
increases from $0.0380$ to $0.1252$ and NorMuon from $0.0419$ to $0.1313$,
corresponding to roughly $3\times$ larger TAIL positional processing. At $4d$,
Muon and NorMuon remain $2.31\times$ and $2.93\times$ above their RoPE
baselines, respectively. The NoPE-induced effect is strongest at $1d$,
suggesting that learned positional compensation is most pronounced when FFN
width is more constrained. AdamW does not follow this trend; at
$4d$, its TAIL amplification decreases from $0.0183$ under RoPE to $0.0133$
under NoPE.

\begin{table}[h]
    \centering
    \setlength{\tabcolsep}{5pt}
    \caption{TAIL positional amplification under RoPE and NoPE for GPT-2 350M.
    Values are layer-averaged $\overline{|\Delta\mathrm{SR}|}$ at the smallest
    and largest FFN widths. Ratios compare NoPE to RoPE.}
    \label{tab:rope_nope_tail_abs}
    \begin{tabular}{l ccc ccc}
    \toprule
    & \multicolumn{3}{c}{$1d$}
    & \multicolumn{3}{c}{$4d$} \\
    \cmidrule(lr){2-4}
    \cmidrule(lr){5-7}
    Optimizer
    & RoPE & NoPE & Ratio
    & RoPE & NoPE & Ratio \\
    \midrule
    AdamW
    & 0.0096 & 0.0212 & 2.21$\times$
    & 0.0183 & 0.0133 & 0.73$\times$ \\
    Muon
    & 0.0380 & 0.1252 & 3.29$\times$
    & 0.0196 & 0.0453 & 2.31$\times$ \\
    NorMuon
    & 0.0419 & 0.1313 & 3.13$\times$
    & 0.0213 & 0.0624 & 2.93$\times$ \\
    \bottomrule
    \end{tabular}
\end{table}

The signed values confirm that this effect corresponds to positional
amplification rather than only a change in magnitude. Under NoPE, Muon and
NorMuon have negative signed $\Delta\mathrm{SR}(\mathrm{TAIL})$ across all
widths, with much larger magnitudes than AdamW. Across settings,
$\mathrm{SR}_{\mathrm{pre}}(\mathrm{TAIL})$ remains high, typically between
$0.95$ and $0.98$, indicating that the largest differences arise in how the FFN
nonlinearity changes position-dependence rather than from different
pre-activation symmetry.

\subsection{NoPE Changes the Depth Localization of Positional Processing}

Table~\ref{tab:rope_nope_peak} reports where TAIL positional amplification is
strongest across layers. Under RoPE, peak TAIL processing for Muon often occurs
in deeper layers (Layer $8,23,23,23$ across the $1d$--$4d$ sweep). Under NoPE,
Muon shifts this peak to the earliest layers (Layer $2,2,1,1$). NorMuon also
shifts earlier under NoPE, but its peak is more distributed across layers (Layer
$3,11,10,2$). AdamW remains concentrated in mid-to-deeper layers under NoPE. Thus,
{\em NoPE changes where positional processing is concentrated in depth, and this
localization differs by optimizer.}

\begin{table}[h]
    \centering
    \small
    \setlength{\tabcolsep}{3.6pt}
    \caption{Peak TAIL positional amplification under RoPE and NoPE for GPT-2
    350M. Each entry reports the most negative TAIL $\Delta\mathrm{SR}$ value,
    with the peak layer in parentheses. Under NoPE, Muon concentrates peak
    positional processing in the earliest layers, while NorMuon shifts earlier
    but remains more distributed. Layer indices are zero-based.}
    \label{tab:rope_nope_peak}
    \resizebox{\textwidth}{!}{%
    \begin{tabular}{l cccc cccc}
    \toprule
    & \multicolumn{4}{c}{RoPE: Peak $\Delta\mathrm{SR}$ (Layer)}
    & \multicolumn{4}{c}{NoPE: Peak $\Delta\mathrm{SR}$ (Layer)} \\
    \cmidrule(lr){2-5}
    \cmidrule(lr){6-9}
    Optimizer
    & $1d$ & $2d$ & $3d$ & $4d$
    & $1d$ & $2d$ & $3d$ & $4d$ \\
    \midrule
    AdamW
    & $-$0.049 (23) & $-$0.039 (7)  & $-$0.046 (23) & $-$0.036 (23)
    & $-$0.257 (15) & $-$0.150 (15) & $-$0.038 (18) & $-$0.127 (17) \\
    Muon
    & $-$0.172 (8)  & $-$0.196 (23) & $-$0.085 (23) & $-$0.160 (23)
    & $-$0.823 (2)  & $-$0.586 (2)  & $-$0.456 (1)  & $-$0.582 (1) \\
    NorMuon
    & $-$0.373 (7)  & $-$0.174 (23) & $-$0.149 (23) & $-$0.109 (6)
    & $-$0.817 (3)  & $-$0.582 (11) & $-$0.374 (10) & $-$0.470 (2) \\
    \bottomrule
    \end{tabular}%
    }
\end{table}

This depth-localization pattern provides a useful lens for the spectral scaling behavior of  RoPE and NoPE
configurations. Muon concentrates strong position-dependent processing very early,
whereas NorMuon produces strong but more distributed processing. This difference
is consistent with the observation that Muon and NorMuon exhibit
different spectral-scaling behavior under NoPE.

\subsection{NoPE Reverses the Frequency Bias of Positional Processing}

We also examine whether the FFN nonlinearity amplifies position-dependence more
for frequent or rare tokens. Since negative $\Delta\mathrm{SR}$ indicates
positional amplification, we report \vspace{-0.6em}
\begin{equation}
    \Delta\mathrm{SR}_{\mathrm{HEAD}}
    -
    \Delta\mathrm{SR}_{\mathrm{TAIL}}.
\end{equation}
Positive values mean that TAIL tokens are amplified more strongly than HEAD
tokens. Under RoPE,
all optimizers and widths have negative values, indicating a HEAD-favored
positional-processing bias. Under NoPE, Muon and NorMuon flip to positive
values across all widths, indicating stronger TAIL positional amplification.
AdamW remains negative under NoPE (Table \ref{tab:head_tail_dsr}).

This reversal shows that removing explicit positional signal changes which
frequency regimes receive stronger FFN-induced positional amplification. Under
NoPE, Muon and NorMuon reorganize toward TAIL-favored positional processing,
whereas AdamW does not show the same reversal. This provides a direct
frequency-level signature of optimizer-dependent positional reorganization
under NoPE.

\begin{table}[h]
    \centering
    \small
    \setlength{\tabcolsep}{4.5pt}
    \caption{HEAD--TAIL positional-processing bias under RoPE and NoPE for
    GPT-2 350M. Values report
    $\Delta\mathrm{SR}_{\mathrm{HEAD}}-\Delta\mathrm{SR}_{\mathrm{TAIL}}$.
    Positive values mean TAIL tokens receive stronger positional amplification.
    Under RoPE, all optimizers are HEAD-biased; under NoPE, Muon and NorMuon
    become TAIL-biased.}
    \label{tab:head_tail_dsr}
    \resizebox{\textwidth}{!}{%
    \begin{tabular}{l cccc cccc}
    \toprule
    & \multicolumn{4}{c}{RoPE}
    & \multicolumn{4}{c}{NoPE} \\
    \cmidrule(lr){2-5}
    \cmidrule(lr){6-9}
    Optimizer
    & $1d$ & $2d$ & $3d$ & $4d$
    & $1d$ & $2d$ & $3d$ & $4d$ \\
    \midrule
    AdamW
    & $-$0.0214 & $-$0.0300 & $-$0.0380 & $-$0.0325
    & $-$0.0212 & $-$0.0428 & $-$0.0238 & $-$0.0060 \\
    Muon
    & $-$0.0064 & $-$0.0173 & $-$0.0207 & $-$0.0151
    & 0.0262 & 0.0256 & 0.0147 & 0.0127 \\
    NorMuon
    & $-$0.0040 & $-$0.0088 & $-$0.0148 & $-$0.0117
    & 0.0438 & 0.0275 & 0.0031 & 0.0032 \\
    \bottomrule
    \end{tabular}%
    }
\end{table}

Hence, above position-dependence analysis provides a structured view of the capacity 
redistribution, as shown in Section~\ref{sec:results:nope}. Precisely,  under NoPE,
Muon increases HEAD hard-rank scaling while decreasing MID and TAIL. 
Whereas NorMuon decreases hard-rank scaling across all token regimes.
The position-dependence analysis provides the explanation; under NoPE, 
Muon and NorMuon amplify TAIL position-dependence,
shift peak processing toward earlier layers, and reverse the HEAD--TAIL
positional-processing bias. However, their localization differs, Muon
concentrates peak processing in the earliest layers, while NorMuon is more
distributed. 

Thus, NoPE changes not only the amount of FFN-induced positional
processing, but also its allocation across depth and token frequency. These
patterns support the broader conclusion that architectural signal changes are
expressed through optimizer-dependent FFN processing rather than as
optimizer-independent perturbations.


\end{document}